\documentclass[journal]{IEEEtran}

  \usepackage{amsmath,amssymb,amsfonts,amsthm,mathrsfs}
  
  \usepackage[ruled, vlined, nofillcomment]{algorithm2e}
  \usepackage{algorithmic}
  
  \SetCommentSty{mycommfont}
  \SetAlgoCaptionSeparator{:}
  
  \newcommand{\T}{^\mathsf{T}}                  
  \newcommand{\tr}{\mathrm{tr}}                 
  \newcommand{\R}{\mathbb{R}}                   
  \providecommand{\norm}[1]{\|#1\|}             
  \newcommand{\diag}[1]{\mathrm{diag}(#1)}      
  \newcommand{\N}{\mathrm{N}}                   
  \newcommand{\GP}{\mathcal{GP}}                
  \newcommand{\dd}{\,\mathrm{d}}                

  \usepackage{bm}
  
  \newcommand{\mbf}[1]{\mathbf{#1}}
  \newcommand{\vect}[1]{\mbf{#1}}
  \newcommand{\vectb}[1]{\bm{#1}}
  \newcommand{\mat}[1]{\mbf{#1}}

  \newcommand{\eg}{\textit{e.g.}\xspace}
  \newcommand{\ie}{\textit{i.e.}\xspace}
  \newcommand{\cf}{\textit{cf.}\xspace}

  \usepackage{xcolor}

  \usepackage{graphicx}
  \graphicspath{ {./}{./fig/} }

  \usepackage{caption}
  \usepackage{subcaption}

  \usepackage{tikz,pgfplots}
  \usetikzlibrary{plotmarks,shapes,arrows}
  \pgfplotsset{compat=newest} 

  \newlength\figureheight
  \newlength\figurewidth

  \usepackage{hyperref}
  \hypersetup{
    bookmarks=true,         
    pdfstartview={FitH},    
    colorlinks=true,        
    linkcolor=black,        
    citecolor=black,        
    filecolor=black,        
    urlcolor=black          
  }

  \usepackage{color}
  \definecolor{manonspurple}{RGB}{148,0,211}

  \definecolor{niklasgreen}{RGB}{0,150,0}

  \usetikzlibrary{external}
  \usepackage[square,numbers,sort&compress]{natbib}

  \usepackage{twoopt,ifthen}
  \renewcommand{\citet}[1]{\citeauthor{#1}~\cite{#1}}
  \renewcommandtwoopt{\citep}[3][][]{\ifthenelse{\equal{#1}{}}{\cite{#3}}{(#1\xspace\cite{#3}\xspace#2)}}

\begin{document}

  \title{Modeling and Interpolation of the Ambient \\ Magnetic Field by Gaussian Processes}

  \author{Arno~Solin, Manon~Kok,~\IEEEmembership{Member,~IEEE}, Niklas~Wahlstr{\"o}m, Thomas~B.~Sch{\"o}n,~\IEEEmembership{Senior~Member,~IEEE,} and~Simo~S{\"a}rkk{\"a},~\IEEEmembership{Senior~Member,~IEEE}
  \thanks{Arno~Solin is with the Department of Computer Science, Aalto University, 02150 Espoo, Finland, and with IndoorAtlas~Ltd., Helsinki, Finland, e-mail: arno.solin@aalto.fi.}
  \thanks{Manon~Kok is with the Department of Engineering, University of Cambridge, CB2~1PZ~Cambridge, United Kingdom, e-mail: mk930@cam.ac.uk.}
\thanks{Niklas Wahlstr\"om and Thomas~B.~Sch\"on are with the Department of Information Technology, Uppsala University, SE-751~05~Uppsala, Sweden, e-mail: \{niklas.wahlstrom, thomas.schon\}@it.uu.se.}
  \thanks{Simo~S\"arkk\"a is with the Department of Electrical Engineering and Automation, Aalto University, 02150 Espoo, Finland, e-mail: simo.sarkka@aalto.fi.}}

\markboth{Accepted to IEEE Transactions on Robotics}%
{Solin \MakeLowercase{\textit{et al.}}: Modeling and Interpolation of the Magnetic Field}

\maketitle

\begin{abstract}
  Anomalies in the ambient magnetic field can be used as features in indoor positioning and navigation. By using Maxwell's equations, we derive and present a Bayesian non-parametric probabilistic modeling approach for interpolation and extrapolation of the magnetic field. We model the magnetic field components jointly by imposing a Gaussian process (GP) prior on the latent scalar potential of the magnetic field. By rewriting the GP model in terms of a Hilbert space representation, we circumvent the computational pitfalls associated with GP modeling and provide a computationally efficient and physically justified modeling tool for the ambient magnetic field. The model allows for sequential updating of the estimate and time-dependent changes in the magnetic field. The model is shown to work well in practice in different applications: we demonstrate mapping of the magnetic field both with an inexpensive Raspberry Pi powered robot and on foot using a standard smartphone.
\end{abstract}


\IEEEpeerreviewmaketitle

\section{Introduction}
\label{sec:introduction}
\begin{figure}[!t]
  \begin{subfigure}[b]{\columnwidth}
    \tikzsetnextfilename{fig01a}

    \setlength{\figurewidth}{\textwidth}
    \setlength{\figureheight}{0.7877\figurewidth}
    \centering\footnotesize%

    \includegraphics{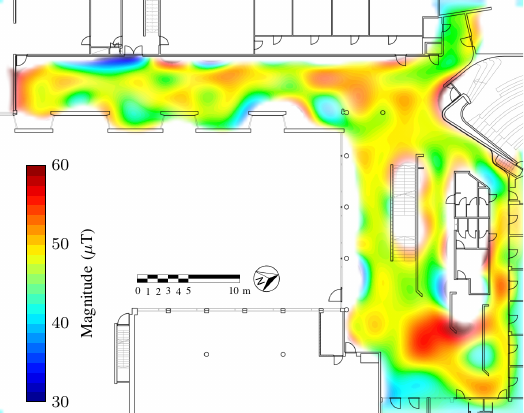}
    \caption{Interpolated magnetic field strength}
  \end{subfigure}
  \hspace*{\fill}  
  \begin{subfigure}[b]{\columnwidth}
    \tikzsetnextfilename{fig01b}
    \centering\footnotesize%
    \includegraphics{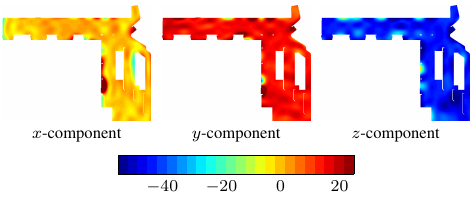}
    \caption{Vector field components}
  \end{subfigure}

  \caption{Interpolated magnetic field of the lobby of a building at the Aalto University campus. The marginal variance (uncertainty) is visualized by the degree of transparency.}
  \label{fig:aalto-estimate}
\end{figure}

Magnetic material causes anomalies in the ambient magnetic field. In indoor environments, large amounts of such magnetic material are present in the structure of buildings and in furniture. Our focus is on building maps of the indoor magnetic field these structures are inducing. These maps are constructed by interpolating three-dimensional magnetic field measurements obtained using magnetometers. An illustration of a map obtained using our proposed method is available in Figure~\ref{fig:aalto-estimate}.

Magnetic maps of indoor environments can be used in indoor positioning and navigation applications \citep[see, \eg,][]{Solin+Sarkka+Kannala+Rahtu:2016}. In these applications, sensors providing position information that is accurate on a short time-scale---but drifts on longer time-horizons---are typically combined with absolute position measurements. For example, data from wheel encoders and inertial sensors---which give accurate position information only on a short time-scale---can be combined with ultra-wideband, Wi-Fi, or optical measurement equipment such as cameras \citep[see, \eg,][]{Woodman:2010,Hol:2011}. The downside of these sources of absolute position is that they typically rely on additional infrastructure or require certain conditions to be fulfilled such as line-of-sight. The advantage of using the magnetic field for positioning is that it can be measured by a small device, without additional infrastructure and without line-of-sight requirements. Furthermore, magnetometers are nowadays present in (almost) any inertial measurement unit (IMU) or smartphone. A requirement for localization using the ambient magnetic field as a source of position information is that accurate maps of the magnetic field can be constructed within reasonable computational complexity which is the focus of this work. This can also be regarded as a step towards simultaneous localization and mapping (SLAM) using magnetic fields in which localization is done while building the map \citep[see, \eg,][]{Leonard+Durrant-Whyte:1991, Durrant-Whyte+Bailey:2006}.

We interpolate the magnetic field using a Bayesian non-parametric approach where prior knowledge about the properties of magnetic fields is incorporated in a Gaussian process (GP) prior. GPs \citep[see, \eg,][]{OHagan:1978,Rasmussen+Williams:2006} are powerful tools for Bayesian non-parametric inference and learning, and they provide a framework for fusing first-principles prior knowledge with quantities of noisy data. This has made them popular tools in signal processing, machine learning, robotics and control~\citep{Cressie:1993, Cressie+Wikle:2011,Deisenroth+Fox+Rasmussen:2015}.

The contributions of this paper are three-fold. First, we model the ambient magnetic field using a Gaussian process prior in which we incorporate \emph{physical knowledge} about the magnetic field. This extends the work by \citet{Wahlstrom+Kok+Schon+Gustafsson:2013} by presenting an approach where the GP prior is a \emph{latent (unobservable) magnetic potential} function. Second, we use a computationally efficient GP implementation that allows us to use the large amounts of data provided by the magnetometer. To circumvent the well-known computational challenges with GPs \citep[see, \eg,][]{Rasmussen+Williams:2006}, we rewrite the model in terms of a Hilbert space representation introduced by \citet{Solin+Sarkka:submitted}. We extend the approach to allow for modeling of the bias caused by the Earth magnetic field. Third, we use this method in combination with the sequential approach introduced in \citet{Sarkka+Solin+Hartikainen:2013}. This allows for \emph{online updating} of the magnetic field estimate. It also opens up the possibility to focus on the spatio-temporal problem in which the magnetic field can change over time, for instance due to furniture being moved around. An extensive evaluation of the proposed method is done using both simulated and empirical data. The simulation study and a small-scale experiment illustrate the feasibility and accuracy of the approach and allow for comparison with other methods. Experiments with a mobile robot and with a hand-held smartphone show the applicability to real-world scenarios.

This paper is structured as follows. The next section covers a survey of existing work, which also provides additional motivation for the approach. Section~\ref{sec:backgroundMagField} provides a brief background of the properties of magnetic fields relevant to this work. The Gaussian process regression model is constructed in Section~\ref{sec:GPprior}, which is then extended to explicit algorithms for batch and sequential estimation in the next section. Section~\ref{sec:experiments} covers the experiments. The experimental results and some additional comments regarding the methodology are discussed at the end of the paper.

\section{Related work}

Spatial properties of the magnetic field have been of interest in a large variety of research domains. For instance, the magnetic field has been extensively studied in geology \citep[see, \eg,][]{Nabighian+Grauch+Hansen+LaFehr+Li+Peirse+Philips+Ruder:2005} but also in magnetospheric physics, geophysics, and astrophysics. In all of these domains, interpolation of the magnetic field is of interest \citep[see, \eg,][for examples of magnetic field interpolation in the respective areas]{
Guillen+Calcagno+Courrioux+Joly+Ledru:2008, Calcagno+Chiles+Courrioux+Guillen:2008, 
Mackay+Marchand+Kabin:2006,Bhattacharyya:1969,Springel:2010}. 

In recent years interest has emerged in using the magnetic field as a source of position information for indoor positioning \citep{Haverinen+Kemppainen:2009}. Feasibility studies have been conducted, focusing both on the time-varying nature of the magnetic field and on the amount of spatial variation in the magnetic field. \citet{Li+Gallagher+Dempster+Rizos:2012} report experiments showing that the magnetic field in a building typically shows large spatial variations and small time variations. This is also supported by the experimental study reported by \citet{Angermann+Frassl+Doniec+Julian+Robertson:2012} in which significant anomalies of the ambient magnetic field are reported. These experiments give confidence that the magnetic field provides sufficient information for localization purposes. However, \citet{Li+Gallagher+Dempster+Rizos:2012} also report significant temporal changes in the magnetic field in the vicinity of mobile magnetic structures, in their case an elevator. 

A number of approaches have been reported on building a map of the ambient magnetic field for indoor localization purposes. \citet{LeGrand+Thrun:2012} propose a method to build a map of the magnetic field by collecting magnetometer data in a grid and linearly interpolating between these points. This map is subsequently used for localization with a particle filter combining magnetometer and accelerometer measurements from a smartphone. \citet{Robertson+Frassl+Angermann+Doniec+Julian+Puyol+Khider+Lichtenstern+Bruno:2013} present a SLAM approach for pedestrian localization using a foot-mounted IMU. They use the magnetic field intensity which they model using spatial binning. \citet{Frassl+Angermann+Lichtenstern+Robertson+Julian+Doniec:2013} discuss the possibility of using more components of the magnetic field (for instance the full three-dimensional measurement vector) in the SLAM approach instead. \citeauthor{Vallivaara+Haverinen+Kemppainen+Roning:2010} \cite{Vallivaara+Haverinen+Kemppainen+Roning:2010, Vallivaara+Haverinen+Kemppainen+Roning:2011} present a SLAM approach for robot localization. They model the ambient magnetic field using a squared exponential GP prior for each of the magnetic field components. \citet{Wahlstrom+Kok+Schon+Gustafsson:2013} incorporate additional physical knowledge by making use of Maxwell's equations resulting in the use of curl- and divergence-free GP priors instead. 

As can be concluded, there exists a wide range of existing literature when it comes to modeling the ambient magnetic field. The amount of information that is used differs between the approaches. For instance, some approaches use full three-dimensional magnetic field vectors while others only use a one-dimensional magnetic field intensity. Furthermore, the amount of physical information that is included differs. In this paper, we build on the approach by \citet{Wahlstrom+Kok+Schon+Gustafsson:2013} and use the full three-dimensional magnetometer measurements. We include physical knowledge in terms of the magnetic field potential.

As discussed above, GPs have frequently been used in modeling and interpolation of magnetic fields. GP regression has also successfully been applied to a wide range of applications \citep[see, \eg,][]{OCallaghan+Ramos:2012, Smith+Posner+Newman:2011, Kim+Kim:2015, Ramos+Ott:2016, Senanayake+Ott+Callaghan+Ramos:2016, Vidal-Calleja+Su+Bruijn+Miro:2014}. Furthermore, it has previously been used for SLAM \citep[see, \eg,][]{Tong+Furgale+Barfoot:2013, Ferris+Hahnel+Fox:2006, Barkby+Williams+Pizarro+Jakuba:2012, Barfoot+Tong+Sarkka:2014, Anderson+Barfoot+Tong+Sarkka:2015}. One of the challenges in using GPs is the computational complexity \citep[see, \eg,][]{Rasmussen+Williams:2006}, which scales cubically with the number of training data points. Considering the high sampling rate of the magnetometer and the fact that each observation contains three values, a large number of measurements is typically available for mapping. Because of these computational challenges, the data in \citet{Wahlstrom+Kok+Schon+Gustafsson:2013} was downsampled.

Attempts to speed up GP inference have spawned a wide range of methods which aimed at bringing GP regression to data-intensive application fields. These methods \citep[see][for a review]{Quinonero-Candela+Rasmussen:2005} typically build upon reducing the rank of the Gram (covariance) matrix and using the matrix inversion lemma to speed up matrix inversion. For stationary covariance functions, the spectral Monte Carlo approximation by \citet{Lazaro-Gredilla+Quinonero-Candela+Rasmussen+Figueiras-Vidal:2010}, the Fourier features by \citet{Hensman+Durrande+Solin:2018}, or the Laplace operator eigenbasis based method introduced by \citet{Solin+Sarkka:submitted} can be employed. For uniformly spaced observations, fast Fourier transforms can provide computational benefits \citep{Paciorek:2007, Fritz+Neuweiler+Nowak:2009}. As will be shown later on in this paper, the Laplace operator approach by \citet{Solin+Sarkka:submitted} falls natural to modeling of the magnetic field in terms of a magnetic field potential.

All approaches on mapping of magnetic fields discussed above assume that the magnetic field is constant over time. However, as shown by \citet{Li+Gallagher+Dempster+Rizos:2012} significant temporal changes in the magnetic field occur in the vicinity of mobile magnetic structures. For GP models evolving in time, spatio-temporal GP models \citep[see, \eg,][]{Cressie+Wikle:2011} can be solved efficiently using Kalman filtering methods \citep{Hartikainen+Sarkka:2010, Reece+Roberts:2010, Osborne:2010, Sarkka+Solin+Hartikainen:2013, Huber:2014}. In this paper, we will take the approach of \citet{Sarkka+Solin+Hartikainen:2013} to compose a spatio-temporal GP prior for the model and solve the inference problem by a sequential Kalman filtering setup. This allows for online estimation of the magnetic field estimate and can be used to allow for time variations in the magnetic field. 

Combining models from physics with GPs has also been studied under the name \emph{Latent force models} by \citeauthor{Alvarez+Luengo+Lawrence:2013} \cite{Alvarez+Lawrence:2009, Alvarez+Luengo+Lawrence:2013}. The connection of these models with spatio-temporal Kalman filtering was studied, for example, in the work by \citeauthor{Hartikainen+Sarkka:2011} (see also \cite{Hartikainen+Sarkka:2011, Sarkka+Hartikainen:2012, Sarkka+Solin+Hartikainen:2013}). However, the Kalman filtering approach itself dates back to \citet{Curtain+Pritchard:1978} in the same way that GPs date back to \citet{OHagan:1978}.

\begin{figure}[t]

  \tikzsetnextfilename{fig02}

  \centering\footnotesize

  \includegraphics{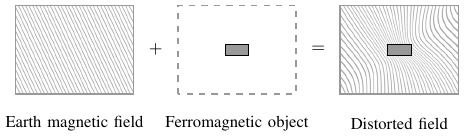}
  \caption{A ferromagnetic object deflects the Earth's magnetic field and introduces  distortions in the field.}

  \label{fig:distorted-field}
\end{figure}

\begin{figure}[!t]

  \tikzsetnextfilename{fig03}

  \setlength{\figurewidth}{.8\columnwidth}
  \setlength{\figureheight}{0.75\figurewidth}
  \centering%

  \includegraphics{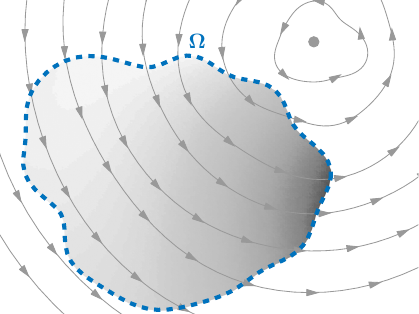}
  \caption{Illustration of a vector field with non-zero curl. The vortex point makes it non-curl-free as the vector field curls around it. However, the subset $\Omega$ excludes the vortex point and the vector field is curl-free in this region. To this region a scalar potential $\varphi$ can be associated, here illustrated with shading.}
  \label{fig:curl}
\end{figure}

\section{The ambient magnetic field}
\label{sec:backgroundMagField}
On a macroscopic scale, magnetic fields are \emph{vector fields}, meaning that at any given location, they have a direction and a strength (magnitude). These properties are familiar from everyday life: the force created by permanent magnets attracting and repelling ferromagnetic materials is used in various utensils, and the compass aligning itself with the direction of the Earth's magnetic field has proved invaluable for mankind during the past centuries. The Earth's magnetic field sets a background for the ambient magnetic field, but deviations caused by the bedrock and anomalies induced by man-built structures deflect the Earth's magnetic field. This makes the magnetic field vary from point to point, see Figure~\ref{fig:distorted-field}.

We describe the magnetic field with a function $\vect{H}(\vect{x})$, where $\vect{H} : \R^3 \to \R^3$. For each point in space $\vect{x}$, there will be an associated magnetic field $\vect{H}(\vect{x})$. Such a vector field can be visualized with field lines, where points in space are associated with arrows.  The principles under which the magnetic field is affected by structures of buildings are well known and governed by the very basic laws of physics \citep[see, \eg,][]{Jackson:1999, Vanderlinde:2004, Griffiths+College:1999}.

In this work, we make use of the fact that the magnetic field $\vect{H}$ is curl-free
\begin{equation} \label{eq:curl_of_H}
  \nabla \times \vect{H} = \vect{0}
\end{equation}
provided that there is no free current (current in wires for example) in the region of interest \citep[see][for more details]{Wahlstrom:2015}. This assumption is valid in most indoor environments where the major source for variations in the ambient field is caused by metallic structures rather than free currents in wires. 

One property of curl-free vector fields is that the line integral along a path $P$ only depends on its starting point $A$ and end point $B$, and not on the route taken
\begin{equation} \label{eq:line_integral}
  \int_{P} \vect{H}(\vect{x}) \cdot \dd\vect{x} = \varphi(A) - \varphi(B),
\end{equation}
where $\varphi : \R^3 \to \R$. This can be rewritten by interpreting $\varphi$ as a scalar potential
\begin{equation} \label{eq:scalar_potential}
  \vect{H} = -\nabla \varphi.
\end{equation}
Figure~\ref{fig:curl} illustrates the curl-free property and the scalar potential. Domain $\Omega$ is curl-free and has an associated scalar potential, while the entire domain is not curl-free due to the vortex point. Note that in a non-curl-free vector field no such scalar potential exists since a line integral around the swirl is non-zero. For the magnetic field $\vect{H}$, the swirl corresponds to a wire of free current pointing perpendicular to the plane, which we assume is not included in the region of interest.

The relation \eqref{eq:scalar_potential} is the key equation that we will exploit in our probabilistic model of the ambient magnetic field. We will choose to model the scalar potential $\varphi$ instead of the magnetic field $\vect{H}$ directly. This implicitly imposes the constraints on the magnetic field that the physics is providing.  This model will be explained in the next section.

\section{Modeling the magnetic field using Gaussian process priors}
\label{sec:GPprior}
In this section, we introduce our approach to modeling and interpolation of the ambient magnetic field. We use a Bayesian non-parametric model in which we use knowledge about the physical properties of the magnetic field as prior information. We tackle the problem of interpolating the magnetic field using Gaussian process (GP) regression. In Section~\ref{subsec:GPbackground} we first give a brief background on GPs. After this, we introduce the problem of modeling the magnetic field in Section~\ref{subsec:interpMag}. A commonly used GP model of the magnetic field is introduced in Section~\ref{subsec:separateModeling}. In Section~\ref{subsec:scalarPot}, we subsequently introduce our proposed method for modeling the magnetic field in which we encode the physical properties that were presented in the previous section.

\subsection{Gaussian process regression}
\label{subsec:GPbackground}
In GP regression \citep{Rasmussen+Williams:2006} the model functions $f(\vect{x})$ are assumed to be realizations from a Gaussian random process prior with a given covariance function $\kappa(\vect{x},\vect{x}')$. Learning amounts to computing the posterior process at some test inputs $\vect{x}_*$ given a set of noisy measurements $y_1,y_2,\dots,y_n$ observed at $\vect{x}_1,\vect{x}_2,\ldots,\vect{x}_n$, respectively. This model is often written in the form
\begin{equation}
\begin{split} \label{eq:generalGP}
  f(\vect{x}) &\sim \GP(0,\kappa(\vect{x},\vect{x}')), \\
          y_i &= f(\vect{x}_i) + \varepsilon_i,
\end{split}
\end{equation}
where the observations $y_i$ are corrupted by Gaussian noise $\varepsilon_i \sim \N(0,\sigma_\text{noise}^2)$, for $i=1,2,\ldots,n$. Because both the prior and the measurement model are Gaussian, the posterior process will also be Gaussian. Hence, the learning problem amounts to computing the conditional means and covariances of the process evaluated at the test inputs.

Prediction of yet unseen process outputs at an input location $\vect{x}_*$ amounts to the following in GP regression: $p(f(\vect{x}_*) \mid \mathcal{D}) = \N(f(\vect{x}_*) \mid \mathbb{E}[f(\vect{x}_*)], \mathbb{V}[f(\vect{x}_*)])$. The conditional mean and variance can be computed in closed-form as \citep[see]{Rasmussen+Williams:2006}
\begin{align}
\begin{split} \label{eq:gp-solution}
  \mathbb{E}[f(\vect{x}_*)] &= 
    \vect{k}_*\T (\vect{K} + \sigma_\text{noise}^2 \vect{I}_n)^{-1} \vect{y}, \\
  \mathbb{V}[f(\vect{x}_*)] &= 
    \kappa(\vect{x}_*,\vect{x}_*) - \vect{k}_*\T 
    (\vect{K} + \sigma_\text{noise}^2 \, \vect{I}_n)^{-1} \vect{k}_*,
\end{split}
\end{align}
where $\vect{K}_{i,j} = \kappa(\vect{x}_i,\vect{x}_j)$, $\vect{k}_*$ is an $n$-dimensional vector with the $i$th entry being $\kappa(\vect{x}_*,\vect{x}_i)$, and $\vect{y}$ is a vector of the $n$ observations. Furthermore, due to Gaussianity, the marginal likelihood (evidence) of the covariance function and noise parameters can also easily be computed, allowing for Bayesian inference of the parameters as well \citep{Rasmussen+Williams:2006}. 

The choice of a specific covariance function encodes the \textit{a priori} knowledge about the underlying process. One of the most commonly used covariance functions, which will also frequently be used in the next sections, is the stationary and isotropic squared exponential (also known as exponentiated quadratic, radial basis function, or Gaussian). Following the standard notation from \citet{Rasmussen+Williams:2006} it is  parametrized as
\begin{equation} \label{eq:cf-se}
  \kappa_\text{SE}(\vect{x},\vect{x}') = 
  \sigma_\text{SE}^2 \, 
    \exp\!\bigg(-\frac{\norm{\vect{x}-\vect{x}'}^2}{2 \, \ell_\text{SE}^2}\bigg),
\end{equation}
where the hyperparameters $\sigma_\text{SE}^2$ and $\ell_\text{SE}$ represent the magnitude scale and  the characteristic length-scale, respectively. These can be learned from data, for instance by maximizing the marginal likelihood. 

\begin{figure*}[!t]

  \begin{subfigure}[b]{0.4\textwidth}

    \tikzsetnextfilename{fig04a}

    \setlength{\figurewidth}{.8\textwidth}
    \setlength{\figureheight}{.8\textwidth}
    \centering\footnotesize%

    \includegraphics{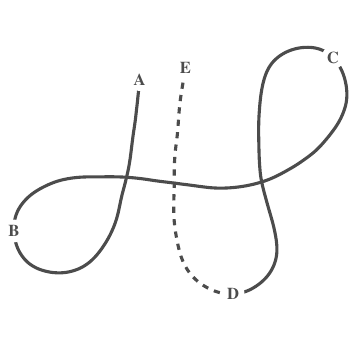}
    \caption{The route}
    \label{fig:gp-demo-a}
  \end{subfigure}
  \begin{subfigure}[b]{0.6\textwidth}

    \tikzsetnextfilename{fig04b}

    \setlength{\figurewidth}{0.9\textwidth}
    \setlength{\figureheight}{0.45\textwidth}
    \centering\footnotesize%

    \includegraphics{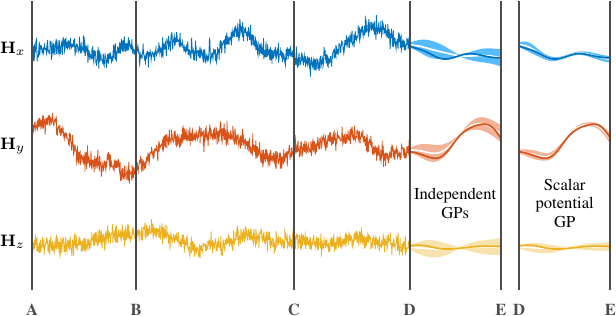}
    \caption{The data and the GP prediction for D--E}
    \label{fig:gp-demo-bc}
  \end{subfigure}
  \caption{A simulated example of the interpolation problem. {(a)}~Training data has been collected along the route A--D, but the magnetic field between D--E is unknown. {(b)}~The noisy observations of the magnetic field between A--D, and GP predictions with 95\% credibility intervals. Both the independent GP modeling approach (with shared hyperparameters) and the scalar potential based curl-free GP approach are visualized. The simulated ground truth is shown by the solid lines.}
  \label{fig:gp-demo} 
\end{figure*}

\subsection{Interpolation of magnetic fields}
\label{subsec:interpMag}
In this work we tackle the problem of interpolating the magnetic field to spatial locations from where we do not have any measurements. In other words, we will tackle the problem of predicting the latent (unobservable noise-free) magnetic field $\vect{f}(\vect{x}_*)$ (such that $\vect{f}: \R^3 \to \R^3$) at some arbitrary location $\vect{x}_*$ given a set of measurements $\mathcal{D} = \{(\vect{x}_i, \vect{y}_i)\}_{i=1}^n$ of the magnetic field. Here, the measurements $\vect{y}$ correspond to the H-field corrupted by i.i.d.\ Gaussian noise. 

Two important things need to be noted with regard to the interpolation of the magnetic field. First, note that the measurements of the magnetic field are \emph{vector-valued} (contrary to the scalar observations in \eqref{eq:generalGP}). This raises the question of how to deal with the different magnetic field components. They can either be treated separately as will be done in Section~\ref{subsec:separateModeling}, or a relation between the different components can be assumed, as is the case in the method we propose in Section~\ref{subsec:scalarPot}. Secondly, note that the function describing the magnetic field is not zero-mean, contrary to the GP model in~\eqref{eq:generalGP}. Instead, its mean lies around a local Earth magnetic field. This depends on the location on the Earth but can also deviate from the Earth's magnetic field in indoor environments due to magnetic material in the structure of the building. The unknown mean can be modeled as an additional part of the covariance function $\kappa(\vect{x},\vect{x}')$ \citep{Rasmussen+Williams:2006}. 

An illustration of GP regression for magnetic fields is provided in Figure~\ref{fig:gp-demo}, where noisy readings of a magnetic field have been collected along route A--D (comprising $\mathcal{D}$), and the magnetic field along route D--E (comprising the prediction locations $\vect{x}_*$) needs to be inferred from the measurements. Each component varies around a local magnetic field due to magnetic material in the vicinity of the sensor. Different interpolation techniques can be used based on different prior knowledge that can be incorporated in the GP. Two different interpolation results are shown: one based on independent modeling of each vector field components (with shared hyperparameters) and another based on associating the GP prior with the scalar potential of the (curl-free) vector field. These are based on the models we will introduce in the coming two sections.

\subsection{Separate modeling of the magnetic field components}
\label{subsec:separateModeling}
The most straightforward approach to GP modeling of vector-valued quantities is to model each of the field components as an independent GP. This approach has been widely applied in existing literature \citep[see, \eg,][]{Vallivaara+Haverinen+Kemppainen+Roning:2010, Vallivaara+Haverinen+Kemppainen+Roning:2011, Kemppainen+Haverinen+Vallivaara+Roning:2011, Jung+Oh+Myung:2015, Ruiz+Olariu:2015}. For each of the three magnetic field components $d \in \{1,2,3\}$, this model can be written as
\begin{equation}
\begin{split} \label{eq:gp-independent}
  f_d(\vect{x}) &\sim \GP(0,
    \kappa_\text{const.}(\vect{x},\vect{x}') + \kappa_\text{SE}(\vect{x},\vect{x}')), \\
  y_{d,i} &= f_d(\vect{x}_i) + \varepsilon_{i,d},
\end{split}
\end{equation}
where the observations $y_{d,i}, i=1,2,\ldots,n$, are corrupted by independent Gaussian noise with variance $\sigma_{\text{noise}}^2$. The non-zero mean of the magnetic field is handled by a constant covariance function (see \cite{Rasmussen+Williams:2006})
\begin{equation} \label{eq:cf-constant}
  \kappa_\text{const.}(\vect{x},\vect{x}') = 
  \sigma_\text{const.}^2,
\end{equation}
where $\sigma_\text{const.}^2$ is a magnitude scale hyperparameter. The small-scale variation in the field is modeled by a squared exponential covariance function~\eqref{eq:cf-se}. Hence, the model~\eqref{eq:gp-independent} encodes the knowledge that the realizations are expected to be smooth functions in space with a constant shift from zero mean.

The model has four hyperparameters: two magnitude scale parameters ($\sigma_\text{const.}^2$ and $\sigma_\text{SE}^2$), a length-scale parameter ($\ell_\text{SE}$), and a noise scale parameter ($\sigma_\text{noise}^2$). Assuming that the components are completely separate, each component has four hyperparameters to learn. The resulting model is flexible, as it does not encode any relation between the vector field components. In practice, this might lead to problems in hyperparameter estimation, with parameter estimates converging to local optima and magnetic field components behaving very differently with respect to each other. Therefore, the hyperparameters are often fixed to reasonable values---instead of learned from data (see, \eg, \cite{Kemppainen+Haverinen+Vallivaara+Roning:2011}).

A more sensible approach for separate, but not completely independent, modeling of the magnetic field measurements, models them as realizations of three independent GP priors with joint learning of the shared hyperparameters \citep[see also][]{Kemppainen+Haverinen+Vallivaara+Roning:2011}. Note that for this model, the covariance in the GP posterior is independent of the outputs $\vect{y}$ and only depends on the input locations $\vect{x}$ (which are shared for all components in $\vect{y}$). Hence, calculating the marginal likelihood only requires inverting a matrix of size $n$ (not $3n$). For this model, the expression for evaluating the log marginal likelihood function for hyperparameter optimization can be written as
\begin{multline}
  \mathcal{L}(\vectb{\theta}) = 
     -\log p(\vect{y} \mid \vectb{\theta}, \mathcal{D}) = 
      \frac{3}{2} \log |\vect{K}_{\vectb{\theta}}+\sigma_\text{noise}^2\,\vect{I}_n| \\
    + \frac{1}{2} \tr \big[ \vect{y}(\vect{K}_{\vectb{\theta}}+\sigma_\text{noise}^2\,\vect{I}_n)^{-1}\vect{y}\T \big]
    + \frac{3n}{2} \log(2\pi),
\end{multline}
where $\vect{y} \in \R^{3 \times n}$ and $\vect{K}_{\vectb{\theta}} \in \R^{n \times n}$.

Figure~\ref{fig:gp-demo-bc} shows the results of predicting the magnetic field behavior along the route D--E for the GP prior modeling the three magnetic field components separately but with joint learning of the shared hyperparameters. The colored patches show the 95\% credibility intervals for the prediction with the mean estimate visualized by the white line. The simulated ground-truth (solid colored line) falls within the shown interval, and the model captures the general shape of the magnetic field variation along the path. The strengths of this model are that it is flexible and that the assumptions are conservative. The weaknesses on the other hand are evident: The model does not incorporate physical knowledge of the magnetic field characteristics. In the next section we will instead explore this knowledge by modeling the magnetic field as derivative measurements of a scalar potential.

\subsection{Modeling the magnetic field as the gradient of a scalar potential}
\label{subsec:scalarPot}
Following our choices in Section~\ref{sec:backgroundMagField}, we assume that the magnetic field $\vect{H}$ can be written as the gradient of a scalar potential $\varphi(\vect{x})$ according to~\eqref{eq:scalar_potential}. Here, $\varphi : \R^3 \to \R$ and $\vect{x} \in \R^3$ is the spatial coordinate. We assume $\varphi(\vect{x})$ to be a realization of a GP prior and the magnetic field measurements $\vect{y}_i \in \R^3$ to be its gradients corrupted by Gaussian noise. This leads to the following model
\begin{equation}
\begin{split} \label{eq:scalar-potential-gp}
  \varphi(\vect{x}) &\sim \GP(0,\kappa_\text{lin.}(\vect{x},\vect{x}') + \kappa_\text{SE}(\vect{x},\vect{x}')), \\
    \vect{y}_i &= -\nabla \varphi(\vect{x}) \big|_{\vect{x}=\vect{x}_i} + \vectb{\varepsilon}_i,
\end{split}
\end{equation}
where $\vectb{\varepsilon}_i \sim \N(\vect{0}, \sigma_\text{noise}^2 \, \mat{I}_3)$, for each observation $i=1,2,\ldots,n$. The squared exponential covariance function~\eqref{eq:cf-se} in~\eqref{eq:scalar-potential-gp} allows us to model the magnetic field anomalies induced by small-scale variations and building structures. The local Earth's magnetic field contributes linearly to the scalar potential as
\begin{equation} \label{eq:cf-linear}
  \kappa_\text{lin.}(\vect{x},\vect{x}') = 
  \sigma_\text{lin.}^2 \, 
    \vect{x}\T\vect{x},
\end{equation}
where $\sigma_\text{lin.}^2$ is the magnitude scale hyperparameter. To simplify the notation in the next sections, we introduce the notation $\vect{f}(\vect{x})$ for the gradient field evaluated at $\vect{x}$. 

Derivative measurements can straightforwardly be used in Gaussian process regression because nabla ($\nabla$) is a linear operator and Gaussianity is preserved under linear operations~\cite{Rasmussen+Williams:2006}. Hence, the model~\eqref{eq:scalar-potential-gp} allows us to learn a map of the magnetic field and make predictions at unseen locations using the methods described in Section~\ref{subsec:GPbackground}. We learn the four hyperparameters of the model---the magnitude scale parameters ($\sigma_\text{lin.}^2$ and $\sigma_\text{SE}^2$), the length-scale parameter ($\ell_\text{SE}$), and the noise scale parameter ($\sigma_\text{noise}^2$)---by maximizing the marginal likelihood.

The model~\eqref{eq:scalar-potential-gp}---where we place a GP prior on the scalar potential $\vect{\varphi}(\vect{x})$ and the magnetometer measurements are derivative measurements---can in fact equivalently be written as a GP prior on the magnetic field $\vect{H}(\vect{x})$ as
\begin{equation}
\begin{split} \label{eq:curl-free-gp}
  \vect{H}(\vect{x}) &\sim \GP(\vect{0},
    \sigma_\text{const.}^2\,\vect{I}_3 + \vect{K}_\text{curl}(\vect{x},\vect{x}')),
\end{split}
\end{equation}
and magnometer measurements being of the standard form~\eqref{eq:generalGP}.
Here, $\vect{K}_\text{curl}(\vect{x},\vect{x}')$ is the so-called curl-free kernel  \citep[see, \eg,][]{Fuselier:2007,Baldassarre+Rosasco+Barla+Verri:2010,Alvarez+Rosasco+Lawrence:2012}. This kernel ensures that any sample drawn from this GP prior obeys the curl-free constraints \eqref{eq:curl_of_H}. The equivalence of~\eqref{eq:scalar-potential-gp} and~\eqref{eq:curl-free-gp} is shown in \citet{Wahlstrom:2015}. In \citet{Jidling+Wahlstrom+Wills+Schon:2017} a more general approach to force GP priors to obey linear constraints is presented. For our work, however, the model formulation through the scalar potential is crucial, as it will enable us to easily extend the model to an approximate form for efficient GP inference in the next section.

The second set of predictions for the route D--E in Figure~\ref{fig:gp-demo-bc} shows the interpolation outcome from the model~\eqref{eq:scalar-potential-gp}. In comparison to the independent GP model, the scalar potential based GP prior provides additional information to the model by tying the vector field components to each other. This improves the estimates in terms of accuracy and makes the 95\% credibility interval more narrow.

\section{Efficient GP modeling of the magnetic field}
Gaussian processes are convenient tools for assigning flexible priors to data---as we saw in the previous section. However, the main problem with the model in the previous section is its high computational cost. The approach scales as $\mathcal{O}(n^3)$ (recall that each observation is three-dimensional, meaning $3n$ becomes large very quickly). This computational complexity renders the approach more or less useless in practice, when the number of observations becomes large (say $n > 1000$).

This is a fundamental restriction associated with the naive formulation of GP models involving the inversion of the covariance matrix. Using special structure of the problem and/or approximative methods, this high computational cost can often be circumvented. In this section we present an approach which both uses the special differential operator structure and projects the model on a set of basis functions characteristic to the covariance function. We first present the method for spatial batch estimation, and then extend it to a temporal dimension as well.

Existing GP methods for mapping and interpolation of the magnetic field have been considering only batch estimation, where the data is first acquired and then processed as a batch. In this section, we aim to extend this to an online method, enabling the GP regression estimate of the magnetic field to be updated when new data is acquired. We denote such a data set as $\mathcal{D}_n = \{(\vect{x}_i, \vect{y}_i) \mid i=1,2,\ldots,n\}$, and thus $\mathcal{D}_i$ denotes all the data that has been observed up to time instance $t_i$.

Considering time as part of the data stream enables us to think of three distinctive setups for estimation of the magnetic field:
\begin{itemize}
  \item \textbf{Batch estimation} of the magnetic field, where the data is first acquired and then the field is estimated at once.
  \item \textbf{Sequential updating} of the field estimate, where we assume all the measurements to be of the same static magnetic field.
  \item \textbf{Spatio-temporal estimation} of the time-dependent magnetic field, where we assume the field to change over time.
\end{itemize}
In the next sections we will present how these scenarios can be combined with the scalar potential based GP scheme without requiring to repeat the batch computations after each sample. 

\subsection{Reduced-rank GP modeling}
A recent paper by \citet{Solin+Sarkka:submitted} presents an approach that is based on a series expansion of stationary covariance functions. The approximation is based on the following truncated series:
\begin{equation} \label{eq:approximation}
  \kappa(\vect{x},\vect{x}')
    \approx \sum_{j=1}^m S(\lambda_j) \, \phi_j(\vect{x})\,\phi_j(\vect{x}'),
\end{equation}
where $S(\cdot)$ is the spectral density of the covariance function $\kappa(\cdot,\cdot)$, $\phi_j(\vect{x})$ is the $j$th eigenfunction of the negative Laplace operator and $\lambda_j^2$ is the corresponding eigenvalue. The efficiency of this approach is based on two properties: (i)~the eigenfunctions are independent of the hyperparameters of the covariance function, and (ii)~for many domains the eigenfunctions and eigenvalues can be solved beforehand in closed-form. Truncating this expansion at degree $m \ll n$ allows the GP regression problem to be solved with a $\mathcal{O}(nm^2)$ and the hyperparameters to be learned with a $\mathcal{O}(m^3)$ time complexity. The memory requirements scale as $\mathcal{O}(n m)$.

Our interest lies in modeling the magnetic field in compact subsets of $\R^3$, allowing us to restrict our interest to domains $\Omega$ comprising three-dimensional cuboids (rectangular boxes) such that $\vect{x} \in [-L_1, L_1] \times [-L_2, L_2] \times [-L_3, L_3] \subset \R^3$ (recall that a stationary covariance function is translation invariant). In this domain, we can solve the eigendecomposition of the Laplace operator subject to Dirichlet boundary conditions
\begin{equation}
\begin{cases}
  -\nabla^2 \phi_j(\vect{x}) = \lambda_j^2 \phi_j(\vect{x}), 
    & \vect{x} \in \Omega, \\
  \phantom{-\nabla^2} \phi_j(\vect{x}) = 0, 
    & \vect{x} \in \partial\Omega.
\end{cases}
\end{equation}
The choice of the domain and boundary conditions is arbitrary, but for regression problems with a stationary covariance function the model reverts back to the prior outside the region of observed data, so the Dirichlet boundary condition does not restrict the modeling if $\Omega$ is chosen suitably. This particular choice of domain and boundary conditions yields the following analytic expression for the basis functions and the corresponding eigenvalues:
\begin{align}
  \phi_j(\vect{x}) =& 
  \prod_{d=1}^3 
    \frac{1}{\sqrt{L_d}} \sin\!\bigg( \frac{\pi n_{j,d} (x_d + L_d)}{2 L_d} \bigg), \\
  \lambda_j^2 =&
  \sum_{d=1}^3 
    \bigg( \frac{\pi n_{j,d}}{2 L_d} \bigg)^2,
\end{align}
where the matrix $\vect{n} \in \R^{m \times 3}$ consists of an index set of permutations of integers $\{1,2,\ldots,m\}$ (\ie, $\{(1, 1,1),(1,1,2),\ldots,(1,2,1),\ldots,(2,1,1), \ldots\}$) . The basis functions are independent of the hyperparameters, and thus only need to be evaluated once.

The Laplace operator eigenbasis approximation method can be combined with the independent GP approach, the independent GPs with shared hyperparameters, and the scalar potential GP approach. Our presentation will be specific to the scalar potential model presented in Section~\ref{subsec:scalarPot}, but a similar setup can be constructed for the other methods as well. The approximation method is even better suited for the scalar potential model because the approximation is based on the eigendecomposition of the Laplace operator. This eigenbasis falls naturally to the problem formulation in which the latent potential field is observed through gradients.

The covariance in our model~\eqref{eq:scalar-potential-gp} consists of a squared exponential and a linear covariance function. The former is stationary and the approach from~\cite{Solin+Sarkka:submitted} can straightforwardly be applied. The latter is not stationary, but there is also no need to approximate it. Consequently, we consider the approximation
\begin{align}
\kappa(\vect{x},\vect{x}') & =  \kappa_\text{lin.}(\vect{x},\vect{x}') + \kappa_\text{SE}(\vect{x},\vect{x}') \notag \\
& \approx \sigma_\text{lin.}^2 \vect{x}\T \vect{x} + \sum_{j=1}^m S_\text{SE}(\lambda_j) \, \phi_j(\vect{x})\,\phi_j(\vect{x}').
\end{align}
Note that adding the linear covariance function, the approximation will no longer revert to zero on the domain boundary, but will instead revert to the scalar potential of the local Earth magnetic field.

The computational benefits come from the approximate eigendecomposition of the Gram (covariance) matrix, $\vect{K}_{i,j} = \kappa(\vect{x}_i,\vect{x}_j)$ \citep[see][for derivations and discussion]{Solin+Sarkka:submitted}. It can now be written out in terms of the basis functions and spectral densities: $\vect{K} \approx \vectb{\Phi}\vectb{\Lambda}\vect{\Phi}\T$. The basis functions, which span the solution, are collected in the matrix $\vectb{\Phi} \in \R^{n \times (3+m)}$, with the following rows
\begin{equation} \label{eq:basis}
  \vectb{\Phi}_i = 
  \begin{pmatrix}
    \vect{x}_i\T, \phi_1(\vect{x}_i), \phi_2(\vect{x}_i), \ldots, \phi_m(\vect{x}_i)
  \end{pmatrix},
\end{equation}
for $i=1,2,\ldots,n$. Accordingly, we define the corresponding measurement model matrix projecting the derivative observations onto the basis functions. Analogously, we define the matrix $\nabla \vectb{\Phi} \in \R^{3n \times (3+m)}$ as the following block-row matrix:
\begin{equation}\label{eq:basis-dx}
  \nabla \vectb{\Phi}_i = 
  \begin{pmatrix}
    \nabla \vect{x}_i\T, \nabla\phi_1(\vect{x}_i), \nabla\phi_2(\vect{x}_i), \ldots, \nabla\phi_m(\vect{x}_i)
  \end{pmatrix},
\end{equation}
for $i=1,2,\ldots,n$. Similarly we define $\vectb{\Phi}_*$ and $\nabla\vectb{\Phi}_*$ as vectors evaluated at the prediction input location $\vect{x}_*$ defined analogously to Equations~\eqref{eq:basis} and \eqref{eq:basis-dx}, respectively. The matrix $\vectb{\Lambda}$ is defined by
\begin{equation}\label{eq:basis-Lambda}
  \vectb{\Lambda} = \diag{\sigma_\text{lin.}^2, \sigma_\text{lin.}^2, \sigma_\text{lin.}^2, S_\text{SE}(\lambda_1), S_\text{SE}(\lambda_2), \ldots, S_\text{SE}(\lambda_m)}.
\end{equation}
For three-dimensional inputs, the spectral density function of the squared exponential covariance function \eqref{eq:cf-se} is given by
\begin{equation} \label{eq:se-spectral-density}
  S_\text{SE}(\omega) = \sigma_\text{SE}^2 \, (2\pi\ell_\text{SE}^2)^{3/2} 
    \exp\!\left(-\frac{\omega^2\ell_\text{SE}^2}{2} \right),
\end{equation}
where the hyperparameters $\sigma_\text{SE}^2$ and $\ell_\text{SE}$ characterize the  spectrum.

\subsection{Batch estimation}
\label{sec:fastApproach}
We first tackle the batch estimation problem which provides the approximative solution to the GP regression problem in Equation~\eqref{eq:gp-solution} for the scalar potential GP. 

Following the derivations of \citet{Solin+Sarkka:submitted}, predictions for interpolation and extrapolation of the magnetic field at yet unseen input locations $\vect{x}_*$ are given by:
\begin{align}
\begin{split} \label{eq:gp-solution-approx}
  \mathbb{E}[\vect{f}(\vect{x}_*)] &\approx 
    \nabla \vectb{\Phi}_*
    ([\nabla\vectb{\Phi}]\T\nabla\vectb{\Phi} + \sigma_\text{noise}^2 \vectb{\Lambda}^{-1})^{-1}
    [\nabla \vectb{\Phi}]\T \mathrm{vec}(\vect{y}), \\
  \mathbb{V}[\vect{f}(\vect{x}_*)] &\approx 
    \sigma_\text{noise}^2 \, \nabla \vectb{\Phi}_* 
    ([\nabla\vectb{\Phi}]\T\nabla\vectb{\Phi} + \sigma_\text{noise}^2 \vectb{\Lambda}^{-1})^{-1}
    [\nabla \vectb{\Phi}_*]\T,
\end{split}
\end{align}
where $\mathrm{vec}(\cdot)$ is the vectorization operator which converts a matrix to a column vector by stacking its columns on top of each other, such that the $3 \times n$ matrix is converted into a vector of size $3n$. The basis functions $\nabla\vectb{\Phi}$ and $\nabla\vectb{\Phi}_*$ need to be evaluated by Equation~\eqref{eq:basis-dx}, and $\vectb{\Lambda}$ by \eqref{eq:basis-Lambda}.

For this model, the expression for evaluating the log marginal likelihood function for hyperparameter optimization can be written as
\begin{multline}\label{eq:likelihood-fast}
  \mathcal{L}(\vectb{\theta})
  = \frac{1}{2} \log |\vect{K}_{\vectb{\theta}}+\sigma_\text{noise}^2\,\vect{I}_{3n}| 
  + \\ \frac{1}{2} \mathrm{vec}(\vect{y})\T(\vect{K}_{\vectb{\theta}} +  \sigma_\text{noise}^2\,\vect{I}_{3n})^{-1}\mathrm{vec}(\vect{y})
  + \frac{3n}{2} \log(2\pi),
\end{multline}
where the quantities can be approximated by
\begin{align}
  &\log |\vect{K}_{\vectb{\theta}}+\sigma_\text{noise}^2\,\vect{I}_{3n}|
   \approx (3n-m)\log \sigma_\text{noise}^2  \nonumber \\
   &\quad + \sum_{j=1}^m [\vectb{\Lambda}_{\vectb{\theta}}]_{j,j} + \log |\sigma_\text{noise}^2\vectb{\Lambda}_{\vectb{\theta}}^{-1} + [\nabla\vectb{\Phi}]\T\nabla\vectb{\Phi}|, \\
  &\mathrm{vec}(\vect{y})\T(\vect{K}_{\vectb{\theta}} + \sigma_\text{noise}^2\,\vect{I}_{3n})^{-1}\mathrm{vec}(\vect{y})  \nonumber \\
  &\quad\approx \frac{1}{\sigma_\text{noise}^2} \big[ \mathrm{vec}(\vect{y})\T\mathrm{vec}(\vect{y}) - \mathrm{vec}(\vect{y})\T\nabla\vectb{\Phi}(\sigma_\text{noise}^2\vectb{\Lambda}_{\vectb{\theta}}^{-1} \nonumber \\
  &\quad + [\nabla\vectb{\Phi}]\T\nabla\vectb{\Phi})^{-1}[\nabla\vectb{\Phi}]\T\mathrm{vec}(\vect{y}) \big],
\end{align}
where the only remaining dependency on the covariance function hyperparameters is in the diagonal matrix $\vect{\Lambda}$ defined through the spectral density in Equation~\eqref{eq:basis-Lambda}. In a software implementation, Cholesky decompositions can be employed for numerical stability in the calculation of determinants and matrix inverses. For optimizing the hyperparameters, gradient based optimizers can be employed \citep[see][for details on deriving the partial derivatives]{Solin+Sarkka:submitted}. 

Algorithm~\ref{alg:batch-estimation} describes the step-by-step workflow for applying these equations in practice. The inputs for the method are the data $\mathcal{D}$ (spatial points and the magnetic field readings), the test points $\vect{x}_*$ to predict at, the domain boundaries $\Omega$, and the approximation degree parameter $m$ (controlling the accuracy of the Hilbert space approximation, see Eq.~\eqref{eq:approximation}). The algorithm returns the marginal mean and variance of the predicted magnetic field at $\vect{x}_*$. The scalar potential could be returned instead by using Equation~\eqref{eq:basis} in step three.

The computational complexity of this method scales as $\mathcal{O}(nm^2)$ for prediction and $\mathcal{O}(m^3)$ for evaluating the marginal likelihood during optimization, which makes it computationally appealing in comparison to the computational complexity of the naive full GP solution which is $\mathcal{O}(n^3)$. The memory requirements scale as $\mathcal{O}(nm)$.

\begin{algorithm}[t]
  \caption{Algorithm for batch estimation of the scalar potential GP magnetic field with the reduced-rank approach.}
  \label{alg:batch-estimation}
  \begin{algorithmic}[1]
    \REQUIRE $\mathcal{D} = \{(\vect{x}_i,\vect{y}_i)\}_{i=1}^n$, $\vect{x}_*$, $\Omega$, $m$.
    \ENSURE $\mathbb{E}[\vect{f}(\vect{x}_*)], \mathbb{V}[\vect{f}(\vect{x}_*)]$.
    \STATE Use Eq.~\eqref{eq:basis-dx} to evaluate the basis functions  $\nabla\vectb{\Phi}$ from $\vect{x}_i$s and $\Omega$.
    \STATE Use Eq.~\eqref{eq:likelihood-fast} to optimize hyperparameters $\vectb{\theta} = \{\sigma_\text{lin.}^2, \sigma_\text{SE}^2, \ell_\text{SE}\vphantom{^2}, \sigma_\text{noise}^2\}$.
    \STATE Use Eq.~\eqref{eq:basis-dx} to evaluate the basis functions $\nabla\vectb{\Phi}_*$ from $\vect{x}_*$s and $\Omega$.
    \STATE Solve the GP regression problem by Eq.~\eqref{eq:gp-solution-approx}.
  \end{algorithmic}
\end{algorithm}

\subsection{Sequential estimation}
\label{sec:sequential}
Many applications require online (sequential) estimates of the magnetic field. The following formulation provides the \emph{same} (within numerical precision) solution as the batch estimation solution \eqref{eq:gp-solution-approx} in the previous section. The inference scheme in the previous section is in practice the solution of a linear Gaussian estimation problem. Sequential solutions for this type of problems have been extensively studied, and this mathematical  formulation is widely known as the \emph{Kalman filter}. The connection between Kalman filtering and Gaussian process regression has recently been studied, for example, in \cite{Osborne:2010, Sarkka+Solin+Hartikainen:2013, Huber:2014}.

Reformulation of a batch problem to a sequential algorithm is discussed (with examples) in the book by \citet{Sarkka:2013}. Following this formulation (and notation to large extent) we may write the following recursion: Initialize $\vectb{\mu}_0 = \vect{0}$ and $\vectb{\Sigma}_0 = \vectb{\Lambda}_{\vectb{\theta}}$ (from the GP prior). For each new observation $i=1,2,\ldots,n$ update the estimate according to
\begin{align}\label{eq:Kalman-update}
\begin{split}
  \vect{S}_i       &= 
    \nabla \vectb{\Phi}_i \vectb{\Sigma}_{i-1} [\nabla \vectb{\Phi}_i]\T +
    \sigma_\text{noise}^2 \, \vect{I}_3, \\
  \vect{K}_i       &= 
    \vectb{\Sigma}_{i-1} [\nabla \vectb{\Phi}_i]\T \vect{S}_i^{-1}, \\
  \vectb{\mu}_i    &= 
    \vectb{\mu}_{i-1} + \vect{K}_i (\vect{y}_i - \nabla \vectb{\Phi}_i \vectb{\mu}_{i-1}), \\
  \vectb{\Sigma}_i &= 
    \vectb{\Sigma}_{i-1} - \vect{K}_i \vect{S}_i \vect{K}_i\T.
\end{split}
\end{align}
This means that for a test input location $\vect{x}_*$ we get predictions for the mean and the variance of the magnetic field which are given by
\begin{equation}\label{eq:projection}
\begin{split}
  \mathbb{E}[\vect{f}(\vect{x}_*) \mid \mathcal{D}_i] &\approx 
    \nabla \vectb{\Phi}_* \, \vectb{\mu}_i, \\
  \mathbb{V}[\vect{f}(\vect{x}_*) \mid \mathcal{D}_i] &\approx 
    \nabla \vectb{\Phi}_* \, \vectb{\Sigma}_i \, [\nabla \vectb{\Phi}_*]\T,
\end{split}
\end{equation}
and conditional on the data observed up to observation $i$. Writing out the conditioning on $\mathcal{D}$ was stripped in the earlier sections for brevity.

Here we do not consider optimization of the hyperparameters $\vectb{\theta}$. The marginal likelihood can be evaluated through the recursion, but in an online setting we suggest optimizing the hyperparameters with some initial batch early in the data collection and then re-optimizing them later on if necessary.

Algorithm~\ref{alg:sequential} presents the scheme of how to apply the equations in practice. The inputs are virtually the same as for the batch algorithm, but now the hyperparameters $\vectb{\theta}$ are considered known. The current estimate can be returned after each iteration loop. The computational complexity of the sequential modeling approach scales as $\mathcal{O}(m^3)$ per update and thus has a total computational complexity of $\mathcal{O}(nm^3)$, but a memory scaling (if intermediate results are not stored) of $\mathcal{O}(m^2)$.

\begin{algorithm}[t]
  \caption{Algorithm for sequential modeling of the scalar potential GP magnetic field estimate. Alternative {(a)} corresponds to the sequential model, and {(b)} to the spatio-temporal modeling approach.}
  \label{alg:sequential}
  \begin{algorithmic}[1]
    \REQUIRE $\mathcal{D}_n$, $\vect{x}_*$, $\Omega$, $m$, $\vectb{\theta}$.
    \ENSURE $\mathbb{E}[\vect{f}(\vect{x}_*)], \mathbb{V}[\vect{f}(\vect{x}_*)]$.

    \STATE Initialize $\vectb{\mu}_0 = \vect{0}$ and $\vectb{\Sigma}_0 = \vectb{\Lambda}_{\vectb{\theta}}$ from Eq.~\eqref{eq:basis-Lambda}.
    \FOR{$i = 1,2, \dots, n$}
      \STATE Evaluate $\nabla\vectb{\Phi}_i$ by Eq.~\eqref{eq:basis-dx} from $\vect{x}_i$.
      \STATE \textbf{(a)} Perform an update by Eq.~\eqref{eq:Kalman-update}. \\ 
             \textbf{(b)} Perform an update by Eqs.~(\ref{eq:prediction-step}--\ref{eq:update-step}).
      \STATE Evaluate the current prediction at $\vect{x}_*$ by Eq.~\eqref{eq:projection}.
    \ENDFOR
  \end{algorithmic}
\end{algorithm}

\subsection{Spatio-temporal modeling}
\label{sec:spatio-temporal}
The sequential model allows for extending the modeling to also track dynamic changes in the magnetic field without virtually any additional computational burden. Let the data $\mathcal{D}_n = \{(t_i,\vect{x}_i,\vect{y}_i)\}_{i=1}^n$ now also comprise a temporal variable $t$ which indicates the time when each observation was acquired.

We present the following spatio-temporal model for tracking changes in the ambient magnetic field. The spatio-temporal GP prior assigned to the scalar potential $\varphi(\vect{x},t)$, depending on both location and time, is defined as follows:
\begin{equation}
\begin{split} \label{eq:spatio-temporal}
  \varphi(\vect{x}, t) &\sim \GP(0,\kappa_\text{lin.}(\vect{x},\vect{x}') + \kappa_\text{SE}(\vect{x},\vect{x}') \kappa_\text{exp}(t,t')),
\end{split}
\end{equation}
where the additional covariance function $\kappa_\text{exp}(t,t')$ defines the prior assumptions of the temporal behavior. This covariance function is defined through
\begin{equation}
  \kappa_\text{exp}(t,t') = \exp\!\bigg(-\frac{|t-t'|}{\ell_\text{time}}\bigg),
\end{equation}
where $\ell_\text{time}$ is a hyperparameter controlling the length-scale of the temporal effects. In the temporal domain, this model is also known as the Ornstein--Uhlenbeck process \citep[see, \eg,][]{Rasmussen+Williams:2006}. The assumption encoded into it is that the phenomenon is continuous but not necessarily differentiable. Therefore it provides a very flexible means of modeling the changing ambient magnetic field. Also note that in Equation~\eqref{eq:spatio-temporal} the temporal effects are only associated with the anomaly component, the bias being tracked as a static component.

Following the derivations in \citet{Hartikainen+Sarkka:2010}, we can write down the dynamic state space model associated with the time evolution of the spatio-temporal GP prior model \eqref{eq:spatio-temporal}
\begin{equation}
\begin{split}
  \vect{A}_i &= \mathrm{blkdiag}(\vect{I}_3, \vect{I}_m \exp(-\Delta t_i / \ell_\text{time})), \\
  \vect{Q}_i &= \mathrm{blkdiag}(\vect{0}_3, \vect{I}_m [1-\exp(-2\Delta t_i / \ell_\text{time})]),
\end{split}
\end{equation}
where $\Delta t_i = t_{i+1}-t_i$ is the time difference between two consecutive samples and $\vect{0}_3$ denotes a $3 \times 3$ zero matrix.

For the time update (Kalman prediction step) we may thus write
\begin{equation}\label{eq:prediction-step}
\begin{split}
  \tilde{\vectb{\mu}}_i &= \vect{A}_{i-1} \vectb{\mu}_{i-1}, \\
  \tilde{\vectb{\Sigma}}_i &= \vect{A}_{i-1} \vectb{\Sigma}_{i-1} \vect{A}_{i-1}\T + \vect{Q}_{i-1},
\end{split}
\end{equation}
and the modified measurement update (Kalman update step)
\begin{equation}\label{eq:update-step}
\begin{split}
  \vect{S}_i       &= 
    \nabla \vectb{\Phi}_i \tilde{\vectb{\Sigma}}_{i} [\nabla \vectb{\Phi}_i]\T +
    \sigma_\text{noise}^2 \, \vect{I}_3, \\
  \vect{K}_i       &= 
    \tilde{\vectb{\Sigma}}_{i} [\nabla \vectb{\Phi}_i]\T \vect{S}_i^{-1}, \\
  \vectb{\mu}_i    &= 
    \tilde{\vectb{\mu}}_{i} + \vect{K}_i (\vect{y}_i - \nabla \vectb{\Phi}_i \tilde{\vectb{\mu}}_{i}), \\
  \vectb{\Sigma}_i &= 
    \tilde{\vectb{\Sigma}}_{i} - \vect{K}_i \vect{S}_i \vect{K}_i\T.
\end{split}
\end{equation}

Algorithm~\ref{alg:sequential} features the workflow of applying the method (option (b)). In practice the only additional input is including the temporal length-scale in $\vectb{\theta}$. The computational costs are the same as for the sequential model.

\begin{figure*}[!t]
  \centering\footnotesize
  \hspace*{\fill}
  \begin{subfigure}[b]{0.46\textwidth}

    \tikzsetnextfilename{tikz-simulatedData-ndata}
    \tikzsetnextfilename{fig05a}

    \setlength{\figurewidth}{0.75\columnwidth}
    \setlength{\figureheight}{0.75\figurewidth}
    \centering\footnotesize%

    \includegraphics{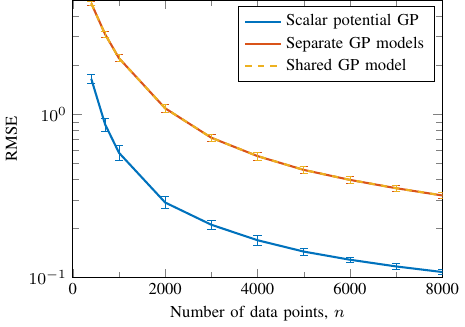}
    \caption{Randomized over data sets}
    \label{fig:simulatedData-ndata}
  \end{subfigure}
  \hspace*{\fill}
  \begin{subfigure}[b]{0.46\textwidth}

    \tikzsetnextfilename{fig05b}

    \setlength{\figurewidth}{0.75\columnwidth}
    \setlength{\figureheight}{0.75\figurewidth}
    \centering\footnotesize%

    \includegraphics{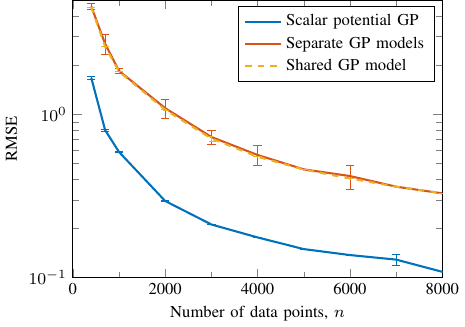}
    \caption{Randomized over initializations}
    \label{fig:simulatedData-init}
  \end{subfigure}\footnotesize
  \hspace*{\fill}
  \caption{The average RMSE (with standard deviations) from 30 Monte Carlo simulations as a function of the number of simulated data points used for training the GP. The constantly lower error for the Hilbert space scalar potential GP in comparison to the separate GP models and shared hyperparameter GP model is explained by the additional prior physical knowledge encoded into the model. In {(b)}, the shared hyperparameter GP shows a slight advantage over the fully independent models.}
  \label{fig:simulated-data}
\end{figure*}

\section{Experiments}
\label{sec:experiments}
The experiments in this paper are split into four parts. We first demonstrate the feasibility of the scalar potential approach with simulated data, where comparison to known ground truth is possible. After this we present a small-scale proof-of-concept demonstration of the approach. The third experiment is concerned with mapping the magnetic field in a building using a handheld smartphone. The final experiment uses an inexpensive mobile robot for online mapping and real-time tracking of the changing magnetic field. The methods were implemented in Matworks \textsc{Matlab}$^{\tiny{\textregistered}}$ on an Apple MacBook Pro (3.1~GHz Intel Core i7, 16~GB RAM).

\subsection{Simulated experiment}
\label{sec:simulated-experiment}
As a first part of our experimental validation of the model, we present a simulation study. This will be used to illustrate our method and to quantify its performance through Monte Carlo simulations. In the simulations, we assume that the magnetic field measurements can indeed be modeled using a scalar potential as argued in Section~\ref{sec:backgroundMagField}. Hence, we simulate the magnetometer data from the GP that models the magnetometer measurements as gradients of a scalar potential field as discussed in Section~\ref{subsec:scalarPot}. Because we are interested in simulating a large number of data points, simulating this data using a full GP approach is computationally very expensive. We therefore simulate the data using the computationally efficient approach described in Section~\ref{sec:fastApproach}. We use a large number of basis functions ($m=4096$). This has been shown to be a good approximation of the true model in~\cite{Solin+Sarkka:submitted}. We will also show that this is a good approximation in Section~\ref{sec:simple-data} based on experimental data. 

The magnetometer is assumed to move in a three-dimensional volume with $x,y,z \in [-0.4,0.4]$ such that $\vect{x} = (x,y,z)$. The training data used for training the GP is randomly uniformly distributed over this volume. A validation data set is used to assess the predictive power of the trained GP. This validation data is a three-dimensional meshgrid over the same volume and consists of $n_\text{val.} = 9261$ positions $\vect{x}_\text{val.}$ with a true magnetic field $\vect{f}_\text{true}(\vect{x}_\text{val.})$. The magnetic field is predicted at these points using the trained GP, leading to $\vect{f}_\text{train.}(\vect{x}_\text{val.})$, after which the quality of the GP solution can be assessed in terms of the root mean square error (RMSE).
We set the domain $\Omega$ in the GP model to $L_1 = L_2 = L_3 = 0.5$ and simulate using the following hyperparameters $\sigma_\text{const.}^2 = 0.3, \sigma_\text{SE}^2 = 1, \ell_\text{SE} = 0.1$, and $\sigma_\text{noise}^2 = 0.04$ (see Sec.~\ref{sec:fastApproach} and~\ref{subsec:scalarPot} for more details on the notation).

We compare our proposed method with two other approaches. The first is the approach by \citeauthor{Vallivaara+Haverinen+Kemppainen+Roning:2010} \cite{Vallivaara+Haverinen+Kemppainen+Roning:2010, Vallivaara+Haverinen+Kemppainen+Roning:2011}, where the magnetic field is modeled using independent GPs for all three components. Each GP consists of a constant and a squared exponential kernel and has its own hyperparameters. The second approach, considered in \citet{Kemppainen+Haverinen+Vallivaara+Roning:2011}, models the magnetic field similarly but with shared hyperparameters. For details, see also Section~\ref{subsec:separateModeling}.

In a first set of Monte Carlo simulations, we analyze the performance of the three different approaches depending on the number of (randomly distributed) training data points, that is in terms of the sparseness of the magnetometer data and the amount of interpolation that is needed for prediction. For all three approaches we use a large number of basis functions ($m = 4096$). They are hence expected to approach the performance of the full GP solution. To exclude problems with local minima in the hyperparameter optimization---which will be the topic of a second set of simulations---the hyperparameter optimization is started in the values used for simulating the data. The results from 30 Monte Carlo simulations are shown in Figure~\ref{fig:simulatedData-ndata}. Naturally, the more data is used for training the GP, the smaller the RMSE becomes. The GP that models the magnetic field measurements as gradients of a scalar potential field, outperforms the other two approaches, independent of the amount of training data used. This can be understood from the fact that this approach incorporates most physical knowledge. 

In a second set of Monte Carlo simulations, the sensitivity to the initialization of the hyperparameter optimization is analyzed for the three different methods. Only one simulated data set is used but the hyperparameter optimization is started in 30 randomly selected sets of hyperparameters $\vectb{\theta}^\text{a}_0$ for a varying length of the training data. The hyperparameters are assumed to lie around the estimates that are obtained using the same optimization strategy as above for 8000 data points. Hence, these sets of hyperparameters $\vectb{\theta}^\text{a}_\text{true}$ are known to results in small RMSE values as depicted in Figure~\ref{fig:simulatedData-ndata}. The superscript `a' on $\vectb{\theta}^\text{a}_\text{true}$ is used to explicitly denote that these sets of hyperparameters actually differ between the three different approaches. For the approach where the magnetic field components are modeled using independent GPs, each component results in a set of hyperparameters $\vectb{\theta}^\text{a}_\text{true}$. For simplicity, in this approach, $\vectb{\theta}^\text{a}_\text{true}$ is chosen to be the mean of these three sets of hyperparameters.

For each of the three approaches, the initial parameters $\vectb{\theta}_0$ are then assumed to deviate from $\vectb{\theta}^\text{a}_\text{true}$ by at most $70\%$ as
\begin{equation}
  \label{eq:simData-randInit}
  \vectb{\theta}_0 = \vectb{\theta}_\text{true}^\text{a} \left[ 1 + 0.7 \, \mathrm{U}(-1,1) \right].
\end{equation}
The Monte Carlo simulation results are depicted in Figure~\ref{fig:simulatedData-init}. As can be seen, the approach which models the three magnetic field components using separate GPs suffers most from local minima. Our proposed model using a scalar potential still outperforms the other two.

\begin{figure}[t]

  \tikzsetnextfilename{fig06}
  \centering\footnotesize%
  \centering\includegraphics{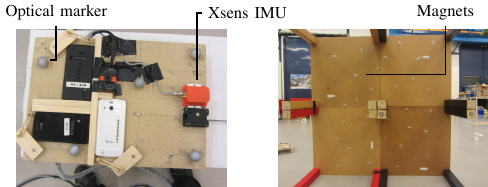}
  \caption{Setup of the experiment: On left the sensor board that was used for collecting the data. 
The right side figure shows the magnetic environment used in the experiment shown from below comprising small magnets to ensure sufficient excitation of the magnetic field. 
}
  \label{fig:expSetup-simple}
\end{figure}

\begin{figure*}[t]

  \setlength{\figurewidth}{0.18\textwidth}

  \pgfplotsset{
    trim axis right,
    colorbar style={ytick style={draw=none}, width=.33cm}
  }
  \begin{subfigure}[b]{0.4\textwidth}
    \centering\footnotesize

    \tikzsetnextfilename{fig07a1}

    \tikzsetnextfilename{fig07a2}

    \includegraphics{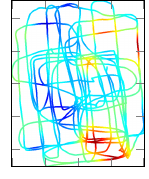}\includegraphics{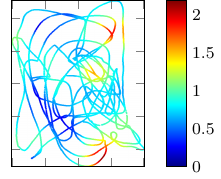}
    \caption{Training data (left) and validation data (right)}
    \label{fig:simpleExample-data-a}
  \end{subfigure}\footnotesize
  \hspace*{\fill}
  \begin{subfigure}[b]{0.50\textwidth}
    \centering\footnotesize

    \tikzsetnextfilename{fig07b}

    \tikzsetnextfilename{fig07b2}

    \includegraphics{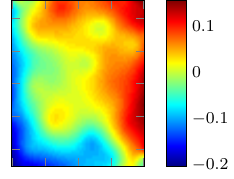}\includegraphics{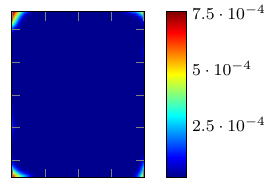}
    \caption{Learned scalar potential (left) and its uncertainty (right)}
    \label{fig:simpleExample-data-b}
  \end{subfigure}
  \hspace*{\fill}
  \\
  \begin{subfigure}[b]{0.7\textwidth}
    \centering\footnotesize

    \tikzsetnextfilename{fig07c1}

    \tikzsetnextfilename{fig07c2}

    \tikzsetnextfilename{fig07c3}

    \includegraphics{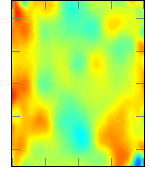}\includegraphics{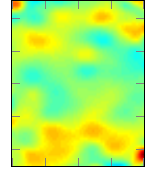}\includegraphics{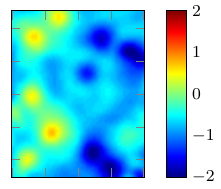}
    \caption{The predicted magnetic field: $\vect{H}_x, \vect{H}_y$, and $\vect{H}_z$}
    \label{fig:simpleExample-data-c}
  \end{subfigure}
  \hspace*{\fill}
  \begin{subfigure}[b]{0.3\textwidth}
    \centering\footnotesize

    \tikzsetnextfilename{fig07d}

    \includegraphics{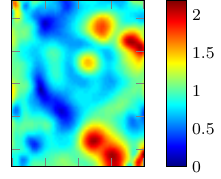}
    \caption{Norm of the magnetic field}
    \label{fig:simpleExample-data-d}
  \end{subfigure}
  \hspace*{\fill}
  \caption{Illustration of the magnetic field data and the results from the GP approach from Algorithm~\ref{alg:batch-estimation} for the experiment discussed in Section~\ref{sec:simple-data}. In all figures, the $y$-axis is $-0.5,\ldots,0.5$~m. The $x$-axis is $-0.4,\ldots,0.4$~m. The units in the field surface plots are arbitrary due to normalization.}
  \label{fig:simpleExample-data}
  \tikzexternalenable
\end{figure*}

\subsection{Empirical proof-of-concept data}
\label{sec:simple-data}
To illustrate our approach using real data, we have performed an experiment where a number of sensors have been moved around in a magnetic environment. The sensor board used is shown in Figure~\ref{fig:expSetup-simple}. We use the magnetometer data from an Xsens  MTi (Xsens Technologies B.V., \mbox{\url{http://www.xsens.com}}). Accurate position and orientation information is obtained using an optical system. These high-accuracy measurements were provided through the use of a Vicon real-time tracking system (Vicon Motion Systems Ltd., UK, \mbox{\url{http://www.vicon.com}}).

We obtain measurements while sliding the sensor board over a configuration of small tables. To ensure sufficient excitation, magnets have been placed in an irregular pattern underneath these tables as shown in Figure~\ref{fig:expSetup-simple}. Two different data sets have been collected. Both consist of approximately three minutes of data sampled at 100~Hz. One data set is used for training, while the second is for validation. The data of both the training and validation data sets are displayed in Figure~\ref{fig:simpleExample-data-a}. To give an impression of the spatial variation of the magnetic field, the magnetic field intensity has been visualized through the colors of the data. Note that the magnetometer is calibrated such that it has a magnitude of one in a local undisturbed magnetic field. 

The magnetometer inside the IMU measures the magnetic field at the different locations. The optical measurements are used for two purposes. First, the positions from the optical system are used as known locations in the GP approach. This is a fairly reasonable assumption due to the high accuracy of the measurements of the optical system. Second, the orientations estimated by the optical system are used to rotate the magnetometer measurements from the magnetometer sensor frame to the lab frame. This rotation of the magnetic field measurements is needed for any of the GP methods discussed in this work. To use the optical and magnetometer data together, they need to be time synchronized. This synchronization is done in post-processing by correlating the angular velocities measured by the optical system and by the gyroscope in the IMU. 

We run Algorithm~\ref{alg:batch-estimation} for the training data set. The domain $\Omega$ in the GP model is set as $L_1 = L_2 = 0.6$ and $L_3 = 0.1$. The actual two-dimensional movement is performed in a rectangle of $80\,\text{cm} \times 100\,\text{cm}$  and is hence well within the domain $\Omega$. The learned scalar potential and its marginal variance (uncertainty) are shown in Figure~\ref{fig:simpleExample-data-b}. Because the training data covers almost the whole displayed area, the learned scalar potential has very low uncertainty. It is also possible to compute the predicted magnetic field which is shown in Figure~\ref{fig:simpleExample-data-c}. For completeness, the intensity of these predicted magnetic field measurements are shown in Figure~\ref{fig:simpleExample-data-d}. Although this quantity is only indirectly related to the outcome of the GP approach, it is frequently used in the remaining sections because of its easy and intuitive visualization. 

\begin{figure}[!t]

  \tikzsetnextfilename{fig08}

  \setlength{\figurewidth}{0.75\columnwidth}
  \setlength{\figureheight}{0.75\figurewidth}
  \centering\footnotesize%

  \includegraphics{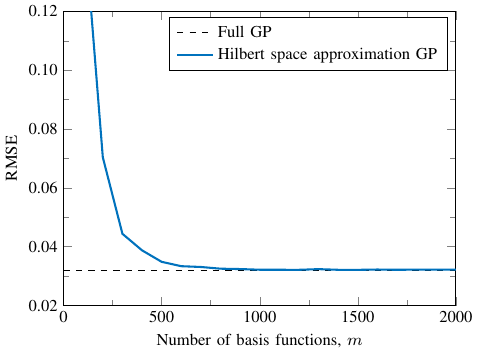}
  \caption{RMSE as a function of the number of basis functions used in Algorithm~\ref{alg:batch-estimation} as compared to a validation data set. The approximative model approaches the full GP approach.
}
  \label{fig:simpleData-nEig}
\end{figure}

By predicting the measurements at the locations of the validation data set, it is also possible to compute the RMSE on the validation data. In Figure~\ref{fig:simpleData-nEig} we visualize the RMSE as a function of the number of basis functions used in Algorithm~\ref{alg:batch-estimation}. We also compare to the RMSE from a full GP approach, that is using the same GP prior but without the Hilbert space approximation scheme to speed up the inference. To allow for comparison with a full GP approach---which suffers from a high computational complexity for large data sets---the data has been downsampled to 5~Hz. As can be seen, already for around $m = 1000$ basis functions, the quality of the estimates from Algorithm~\ref{alg:batch-estimation} approaches that of the full GP approach.

\subsection{Mapping the magnetic field in a building}
\label{sec:experiment-aalto}
\begin{figure}[!t]

  \tikzsetnextfilename{fig09}

  \centering\footnotesize%

  \includegraphics{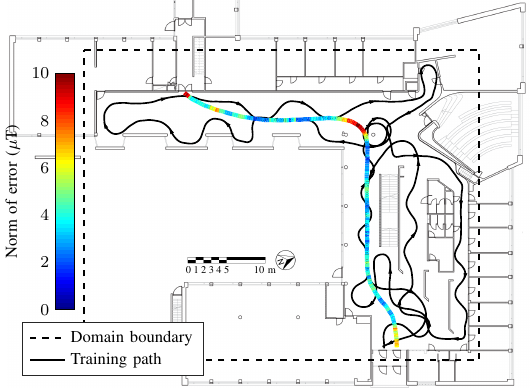}
  \caption{A training (black) and validation (colored) free-walking path that was used in the experiment. Trajectories were collected by a mobile phone, and the magnetometer data was corrected for gravitation direction and heading using the inertial sensors in the device. The domain boundaries for the reduced-rank method are shown by the dashed line.}
  \label{fig:aalto-paths}
\end{figure}

The third experiment was concerned with estimating a map of the magnetic environment inside a building by only using a smartphone for the data collection. The mapped venue is located on the Aalto University campus, and a floor plan sketch is shown in Figure~\ref{fig:aalto-paths}. For practical reasons, we limited our interest to the lobby which is approximately 600~m\textsuperscript{2} in size.

For the measurements, we used an Apple iPhone~4 and its built-in 9-dof IMU (3-axis AKM AK8975 magnetometer). All sensors were sampled at 50~Hz, and the data was streamed online to a laptop computer for processing and storing. The phone was held at waist-height and pointed towards the heading direction.

For reconstructing the walking path and phone orientation, we used a pedestrian dead-reckoning (PDR) approach developed at IndoorAtlas (IndoorAtlas Ltd., Finland, \mbox{\url{http://www.indooratlas.com}}), where only accelerometer and gyroscope readings were used---the path reconstruction thus being fully independent of the magnetometer readings. The alignment to the map and drift correction were inferred from a set of fixed points along the path during acquisition. Two sample paths are shown in Figure~\ref{fig:aalto-paths}: one of the three training paths with similar routes and the validation path. The reconstructed paths were visually checked to match the `true' walking paths.\sloppy

We covered the walkable area in the lobby with three walking paths following the same route in each of them (see Fig.~\ref{fig:aalto-paths}). The reconstructed paths were approximately 242, 253, and 302~meters long, respectively. The number of magnetometer data samples acquired along the paths were 9868, 10500, and 12335. Prior to each acquisition, the phone magnetometer was calibrated by a standard spherical calibration approach. The combined size of the training data set was $n = 32703$. For validation, we collected a walking path passing through the venue (length 54~m, $n=2340$).

We considered a batch interpolation problem of creating a magnetic map of the lobby. The map was assumed static over time, and we applied Algorithm~\ref{alg:batch-estimation} to the training data with $m=1024$ basis functions. The optimized hyperparameters were $\sigma_{\text{lin.}}^2 \approx 575\,(\mu\text{T})^2$, $\sigma_{\text{SE}}^2/\ell_{\text{SE}}^2 \approx 373\,(\mu\text{T})^2$, $\ell_{\text{SE}} \approx 1.87\,\text{m}$, and $\sigma_{\text{noise}}^2 \approx 5.53\,(\mu\text{T})^2$. The coefficients of the inferred linear (bias) model corresponding to the linear covariance function was $(-1.095, 12.995, -41.119)$.

Figure~\ref{fig:aalto-estimate} shows the interpolated magnetic field magnitude ($\norm{\vect{f}}$) and the vector field components. The component-wise maximum of the associated marginal standard deviation fields is visualized by the degree of transparency (with a standard deviation of $5\,\mu\text{T}$ being fully transparent). The overall shape of the estimate agreed even when the model was trained separately with each of the training paths. To most part, the strong fluctuations in the magnetic field are located near walls or other structures in the building. The strong magnetic field in the open area in the lower right part of the floorplan was identified to most likely be due to a large supporting structure on the lower floor-level. We also used the model for predicting the measurements along the validation path (see Fig.~\ref{fig:aalto-paths}). The component-wise RMSEs were $(2.35\,\mu\text{T}, 3.05\,\mu\text{T}, 2.71\,\mu\text{T})$ and mean absolute errors $(1.72\,\mu\text{T}, 2.42\,\mu\text{T}, 2.03\,\mu\text{T})$. Figure~\ref{fig:aalto-paths} shows the norm of the error along the validation path. The measurement noise level of the magnetometer is in the magnitude of $1\,\mu\text{T}$ and the uncertainty in the PDR estimate contributes to the remaining variance.

\subsection{Online mapping}
\label{sec:experiment-online}
Finally, we demonstrate the power of sequential updating and time-dependent magnetic field estimation. The mapping was performed by a lightweight and inexpensive mobile robot equipped with a magnetometer, and the task was to obtain an estimate of the magnetic environment of an indoor space by re-calculating the estimate in an online fashion. In the second part of the experiment the  magnetic environment was abruptly changed during the experiment, and the aim was to catch this phenomenon by spatio-temporal modeling.

We used a robot for collecting the data. The robot was built on a DiddyBorg (PiBorg Inc., UK, \mbox{\url{http://www.piborg.org}}) robotics board, controlled by a Raspberry Pi~2 (model~B) single-board computer (Raspberry Pi Foundation, UK, \mbox{\url{http://www.raspberrypi.org}}). For this example, we controlled the robot over Bluetooth with a joystick.

The robot was equipped with a 9-dof MPU-9150 Invensense IMU unit that was sampled at 50~Hz.  The data were collected and stored internally on the Raspberry Pi. For additional validation, a Trivisio Colibri wireless IMU (TRIVISIO Prototyping GmbH, \mbox{\url{http://www.trivisio.com}}), sampled at 100~Hz, and a Google Nexus~5 smartphone (AKM AK8963 3-axis magnetometer), sampled at 50~Hz, were also mounted on the robot for checking the quality of the Invensense IMU data and ensure the repeatability of the experiment. To reduce disturbances caused by the robot, the sensors were mounted on an approximately 20~cm thick layer of Styrofoam. During post-processing the data, the sensor positions and alignments on the robot were corrected for.

\begin{figure}[!t]

  \tikzsetnextfilename{fig10}
  \centering\footnotesize%
  \includegraphics{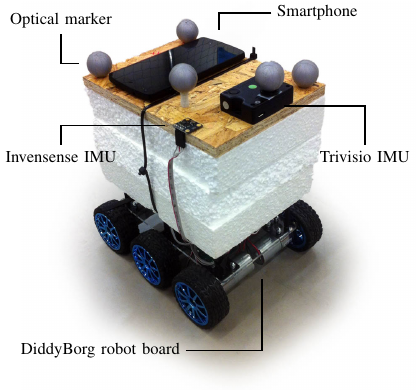}
  \caption{The robot was built on a DiddyBorg robotics board and controlled by a Raspberry Pi single-board computer. The three sensors (Invensense, Trivisio, and smartphone) providing magnetometer readings were mounted on the top. The reference locations were provided by a Vicon optical tracking system.}
  \label{fig:robot-sketch}
\end{figure}

\begin{figure*}[!t]
  \centering
  \begin{subfigure}[b]{0.65\textwidth}

    \tikzsetnextfilename{fig11a}
    \centering\footnotesize
    \includegraphics{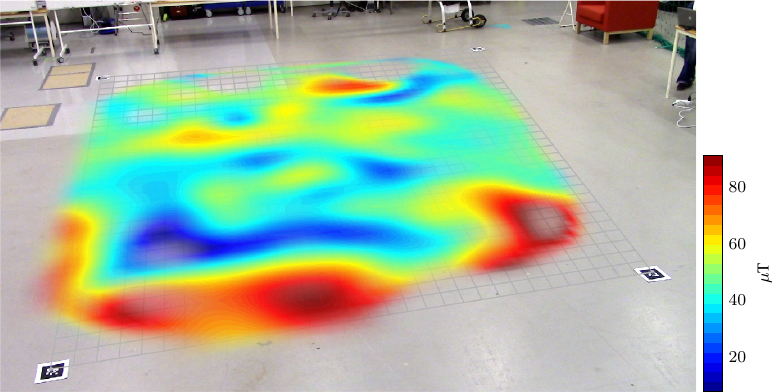}
    \caption{Final interpolated magnitude map}
    \label{fig:robot-final-mag}
  \end{subfigure}
  \hspace*{\fill}%
  \begin{subfigure}[b]{0.28\textwidth}

    \tikzsetnextfilename{fig11b}
    \centering\footnotesize
    \includegraphics{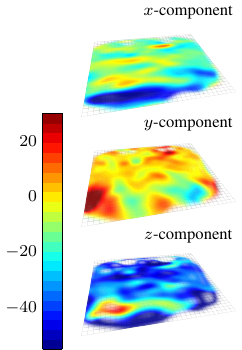}
    \caption{Vector field}
    \label{fig:robot-final-xyz}
  \end{subfigure}
  \\
  \vspace*{12pt} %
  \begin{subfigure}[b]{0.18\textwidth}
    \includegraphics[width=2.6cm]{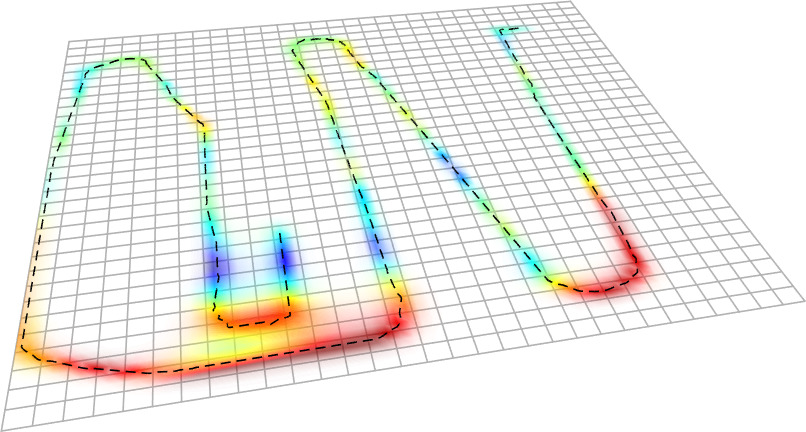}
    \caption{Snapshot~1}
    \label{fig:robot-updates-1}
  \end{subfigure}
  \hfill
  \begin{subfigure}[b]{0.18\textwidth}
    \includegraphics[width=2.6cm]{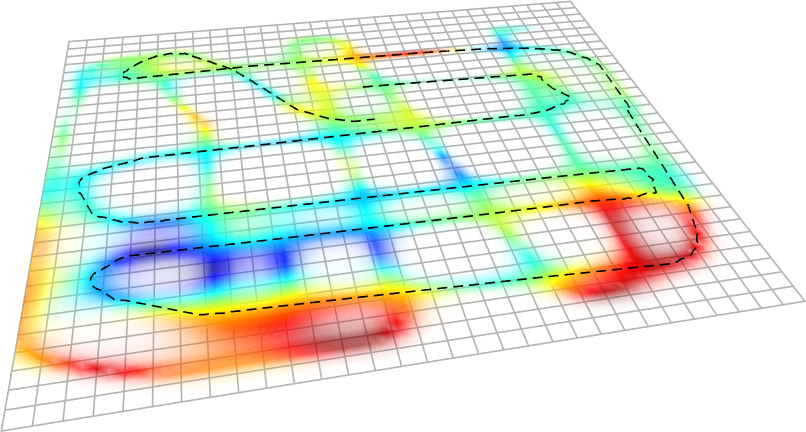}
    \caption{Snapshot~2}
    \label{fig:robot-updates-2}
  \end{subfigure}
  \hfill
  \begin{subfigure}[b]{0.18\textwidth}
    \includegraphics[width=2.6cm]{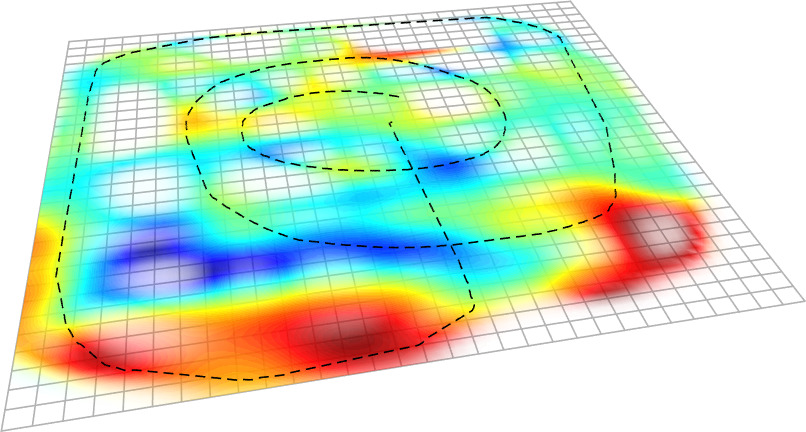}
    \caption{Snapshot~3}
    \label{fig:robot-updates-3}
  \end{subfigure}
  \hfill
  \begin{subfigure}[b]{0.18\textwidth}
    \includegraphics[width=2.6cm]{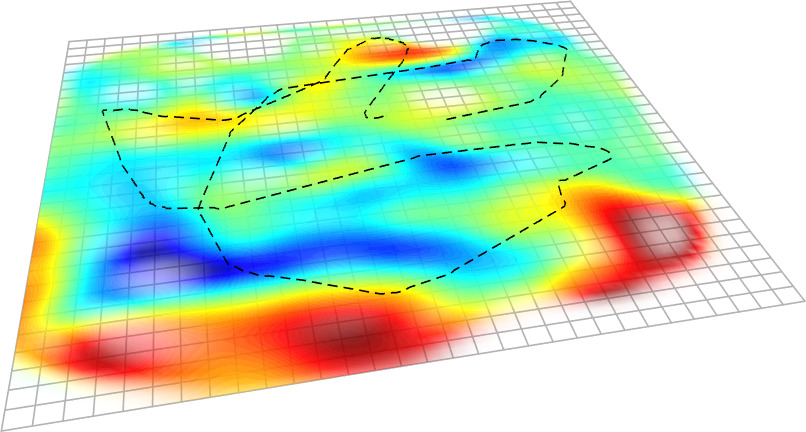}
    \caption{Snapshot~4}
    \label{fig:robot-updates-4}
  \end{subfigure}
  \hfill
  \begin{subfigure}[b]{0.18\textwidth}
    \includegraphics[width=2.6cm]{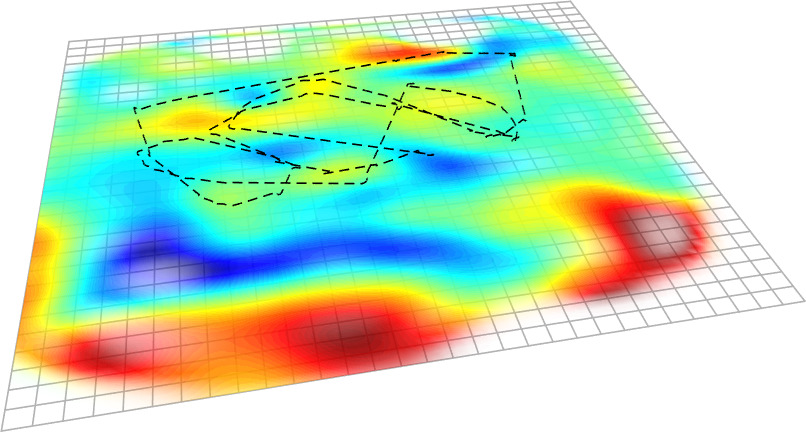}
    \caption{Snapshot~5}
    \label{fig:robot-updates-5}
  \end{subfigure}
  \caption{The GP interpolation task the robot was faced with. The final interpolation outcome of the magnitude field is shown in {(a)}, and the different vector field components are shown in {(b)}. Snapshots along the temporally updating field estimate are shown in {(c--g)} together with the path travelled since the previous update. The marginal variance (uncertainty) is visualized by the degree of transparency.}
  \label{fig:robot-experiment} 
\end{figure*}

High-accuracy location and orientation reference measurements were provided through the use of a Vicon real-time tracking system. The  location measurements could alternatively be recovered by odometry and heading information provided by the robot, but the interest in this experiment was rather to focus on the interpolation of the magnetic field, not the path estimation.

The task was to map the magnetic field inside a marked region roughly 6~m $\times$ 6~m in size. The size of the region was limited by the field of view of the Vicon system. The magnetometers were calibrated in the beginning of the measurement session by rotating the robot around all of its axes. A standard spherical calibration approach was used. Due to limits in acquisition length of the Vicon system, we captured the data in parts, each roughly three minutes in length. The magnetic environment remained unchanged for the first five data sets (paths shown in Figs.~\ref{fig:robot-updates-1}--\ref{fig:robot-updates-5}), and later on changes in the field were initiated by bringing in large metallic toolbox shelves.

In the first part of the experiment, for interpolating the magnetic field we used the sequential reduced-rank scalar-potential approach presented in Algorithm~\ref{alg:sequential}. We assumed the magnetic environment to be stationary, and performed sequential updates in an online fashion. For practical reasons the calculations were done off-line, but the algorithm is fast enough for running in real-time online estimation. The rank of the approximation was fixed to $m=1024$.

The length-scale, magnitude, and noise variance hyperparameters were learned from the first two data sets ($n=17980$ vector valued observations) by maximizing with respect to marginal likelihood (see Alg.~\ref{alg:batch-estimation}). The obtained values were $\ell_\text{SE} \approx 0.32~\text{m}$, $\sigma_\text{SE}^2/\ell_\text{SE}^2 \approx 287\,(\mu\text{T})^2$ and $\sigma_\text{noise}^2 \approx 3.27\,(\mu\text{T})^2$. The linear model magnitude scale parameter was fixed to $\sigma_\text{lin.}^2 = 500\,(\mu\text{T})^2$. The noise model is not only capturing the sensor measurement noise, but the entire mismatch between the data and the model. This explains the rather large noise variance. We also checked, that the hyperparameter estimates remained stable when optimized using the rest of the data. 

Figure~\ref{fig:robot-experiment} shows the results for the static magnetic field experiment. The estimate was updated continuously five times a second, and we show five snapshots of the evolution of the magnetic field estimate in Figures~\ref{fig:robot-updates-1}--\ref{fig:robot-updates-5} (vector field magnitude shown in figures). These snapshots also show the path travelled since the previous snapshot. The alpha channel acts as a proxy for uncertainty; the marginal variance of the estimate is giving the degree of transparency. The final magnitude estimate---after iterating through all the $n=43029$ observations---is shown in Figure~\ref{fig:robot-final-mag} together with the vector field components in Figure~\ref{fig:robot-final-xyz}. 

The frontal part of the mapped region shows strong magnetic activity, whereas the parts further back do not show as strong fields. Inspection of the venue suggested metallic pipelines or structures in the floor to blame (or thank) for these features. In this particular case most parts of the effect is seen in the $x$-component. We repeated the reconstruction with data collected from the Trivisio and smartphone sensors, and the results and conclusions remained unchanged. As a supplementary file to this paper, there is a video\footnote{The supplementary video is available on YouTube: \url{https://www.youtube.com/watch?v=enlMiUqPVJo}} demonstrating the online operation which has been sped-up $50\times$.

\begin{figure}[!t]

  \begin{subfigure}[b]{\columnwidth}

    \tikzsetnextfilename{fig12a}

    \pgfplotsset{clip=false,enlarge x limits=false}

    \setlength{\figurewidth}{0.75\textwidth}
    \setlength{\figureheight}{0.75\figurewidth}
    \centering\footnotesize%

    \includegraphics{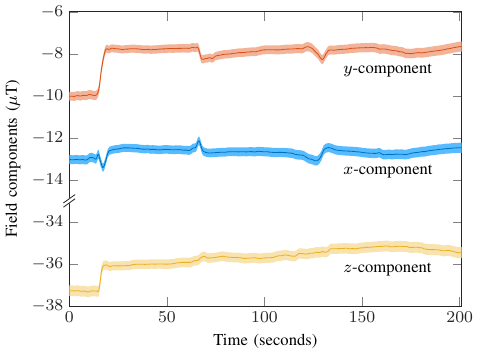}
    \caption{Magnetic field estimate at a fixed location}
    \label{fig:time-evolution}
  \end{subfigure}
  \\ \\
  \begin{subfigure}[b]{0.3\columnwidth}
    \includegraphics[width=\textwidth]{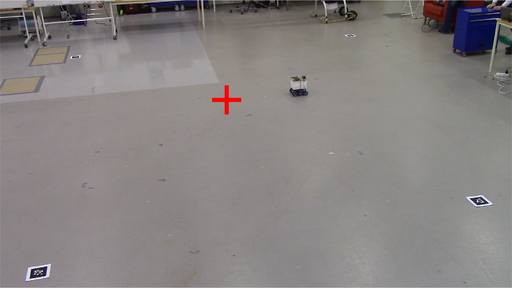}
    \caption{$t=0$ s}
    \label{fig:robot-environment-a}
  \end{subfigure}
  \hfill
  \begin{subfigure}[b]{0.3\columnwidth}
    \includegraphics[width=\textwidth]{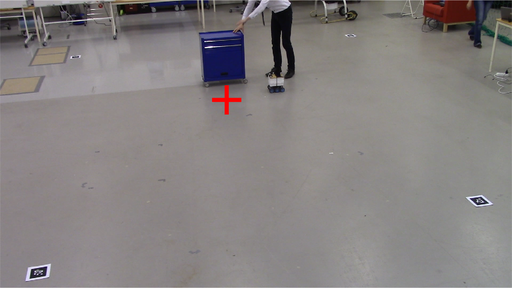}
    \caption{$t=6$ s}
    \label{fig:robot-environment-b}
  \end{subfigure}
  \hfill
  \begin{subfigure}[b]{0.3\columnwidth}
    \includegraphics[width=\textwidth]{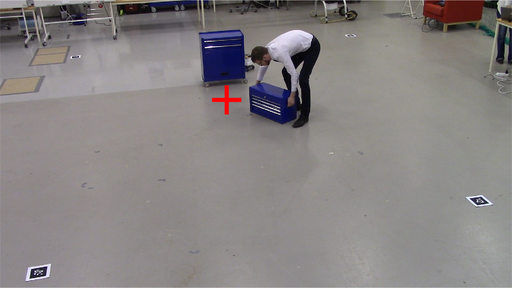}
    \caption{$t=27$ s}
    \label{fig:robot-environment-c}
  \end{subfigure}
  \caption{{(a)}~Evolution of the magnetic field at one spatial location over a time-course of 200 seconds. {(b--d)}~The blue metallic toolboxes are brought in at the beginning of the experiment. The abrupt changes in the field estimate corresponds to time instances when the robot has passed the toolboxes and gained information about the changed environment.}
  \label{fig:dynamic}
\end{figure}

The last part of the experiment was dedicated to dynamical (time-dependent) modeling of the magnetic field.  We used all the data from the first part of the experiment to train a sequential model and used that as the starting point for changing the field ($t=0$). During acquisition of data while the robot was driving around, we brought in two metallic toolbox shelves: first a larger toolbox shelf on wheels (Fig.~\ref{fig:robot-environment-b}) and then a smaller box (Fig.~\ref{fig:robot-environment-c}). We acquired altogether some $300\,$s of data ($n=15513$) of the changed environment.

For encoding the assumptions of a changing magnetic field, we used Algorithm~\ref{alg:sequential}. The additional hyperparameter controlling the temporal scale was fixed to $\ell_\text{time} = 1\,$hour, thus encoding an assumption of slow local changes. This choice is not restrictive, because the data is very informative about the abrupt changes.

Figure~\ref{fig:time-evolution} shows the evolution of the magnetic field components for one fixed location (indicated by a cross marker in the figures). The two toolboxes induce clear changes in the local anomaly field, but the effects are restricted to the immediate vicinity of the boxes. Thus the spatio-temporal model only gains information about the changed field, when the robot passes by the location of interest. This effect is clearly visible around $t=20\,$s, and later on around $t=70\,$s and $t=130\,$s. After this the estimate stabilizes and only drifts around for the remaining time. Even though the changes in the field components appear clear in Figure~\ref{fig:time-evolution}, they are only around $2\,\mu\text{T}$ and thus only account for a variation of about 2\% in the scale of the entire field visualized in Figure~\ref{fig:robot-final-xyz}. Inducing more noticeable changes in the magnetic field would require moving around larger structures (say an elevator). Yet, even changes this small can be tracked by the modeling approach.

\section{Discussion}
\label{sec:discussion}

In recent years, interest has emerged in mapping of the magnetic field by Gaussian processes for robot and pedestrian localization. This paper has aimed at presenting a new efficient method for mapping, but also at providing a study of best practices in using GPs in this context. Thus, we went through three different approaches for formulating GP priors for the magnetic field (independent, shared hyperparameters, and curl-free/scalar potential). These three models differ in the amount of prior knowledge encoded in the model, and the more information available in the prior, the better the interpolation and extrapolation capabilities in the model---as long as the data agrees with the assumptions. This was also demonstrated in Figure~\ref{fig:gp-demo} and in Section~\ref{sec:simulated-experiment}.

The methods presented in this paper are related to the use of Gaussian random field priors in inverse problems \citep{Tarantola:2004, Kaipio+Somersalo:2005}. This connection has been explored from various points of views \citep[\cf][]{Sarkka:2011b}. However, the machine learning \citep{Rasmussen+Williams:2006} way of interpreting the GP priors indeed brings something new on the table---we are explicitly modeling the uncertainty in the field by using a stochastic model, which has interesting philosophical implications. The formulation of the prior through a GP covariance function provides both an intuitive and theoretically justified way of encoding the information.

The scalar potential approach is not the only way to build the model. It would also be possible to include disturbance in the model, which would model the effect of free currents in the area or its boundaries. Furthermore, more complicated assumptions of the temporal time-changing behavior of the magnetic field could be included in the temporal covariance function. For example, various degrees of smoothness or periodicity could be included in the framework.

This paper has been considering the `M' (mapping) part in SLAM. The `L' (localization) part based on these maps is presented in \citet{Solin+Sarkka+Kannala+Rahtu:2016}. The online mapping scheme presented in Algorithm~\ref{alg:sequential} opens up for possibilities for simultaneously building the map and localization within the map. As seen in the experiments presented both in this work and in \citet{Solin+Sarkka+Kannala+Rahtu:2016}, this appears feasible and provides an interesting direction for further research.

\section{Conclusion}
Small variations in the magnetic field can be used as inputs in various positioning and tracking applications. In this paper, we introduced an effective and practically feasible approach for mapping these anomalies. We encoded prior knowledge from Maxwell's equations for magnetostatics into a Bayesian non-parametric probabilistic model for interpolation and extrapolation of the magnetic field.

The magnetic vector field components were modeled jointly by a Gaussian process model, where the prior was associated directly with a latent scalar potential function. This ensures the field to be curl-free---a justified assumption in free spaces. This assumption couples the vector field components and additionally encodes the assumption of a baseline field with smooth small-scale variations. We also presented connections to existing formulations for vector-valued Gaussian process models. 

In addition to constructing the model, we also presented a novel and computationally efficient inference scheme for interpolation and extrapolation using it. We built upon a Laplace operator eigenbasis approach, which falls natural to the formulation of the model. The inference scheme ensures a linear computational complexity with respect to the number of observations of the magnetic field. We also extended the method to an online approach with sequential updating of the estimate and time-dependent changes in the magnetic field.

We presented four experiments demonstrating the feasibility and practicality of the methods. A simulated experiment showed the benefit of including additional knowledge from physics into the model, and a simple proof-of-concept example demonstrated the strength of the approximation scheme in solving the model. Two real-world use cases were also considered: we mapped the magnetic field in a building on foot using a smartphone, and demonstrated online mapping using a wheeled robot.

\section*{Acknowledgements}
This work was supported by grants from the Academy of Finland (266940, 273475, 308640), by CADICS, a Linnaeus Center, and by the project Probabilistic modelling of dynamical systems (contract number: 621-2013-5524), funded both by the Swedish Research Council (VR), by the Swedish Foundation for Strategic Research under the project Cooperative Localization and by the EPSRC grant \emph{Autonomous behaviour and learning in an uncertain world} (Grant number: EP/J012300/1). 

We are grateful for the help and equipment provided by the UAS Technologies Lab, Artificial Intelligence and Integrated Computer Systems Division (AIICS) at the Department of Computer and Information Science (IDA), Link\"oping University, Sweden.
We also acknowledge the computational resources provided by the Aalto Science-IT project.
Finally, we would like to thank IndoorAtlas Ltd.\ for providing expertise and for lending us equipment for the measurements.

{\small
\bibliographystyle{IEEEtranN}
\bibliography{bibliography}

\begin{thebibliography}{67}
\providecommand{\natexlab}[1]{#1}
\providecommand{\url}[1]{#1}
\csname url@samestyle\endcsname
\providecommand{\newblock}{\relax}
\providecommand{\bibinfo}[2]{#2}
\providecommand{\BIBentrySTDinterwordspacing}{\spaceskip=0pt\relax}
\providecommand{\BIBentryALTinterwordstretchfactor}{4}
\providecommand{\BIBentryALTinterwordspacing}{\spaceskip=\fontdimen2\font plus
\BIBentryALTinterwordstretchfactor\fontdimen3\font minus
  \fontdimen4\font\relax}
\providecommand{\BIBforeignlanguage}[2]{{%
\expandafter\ifx\csname l@#1\endcsname\relax
\typeout{** WARNING: IEEEtranN.bst: No hyphenation pattern has been}%
\typeout{** loaded for the language `#1'. Using the pattern for}%
\typeout{** the default language instead.}%
\else
\language=\csname l@#1\endcsname
\fi
#2}}
\providecommand{\BIBdecl}{\relax}
\BIBdecl

\bibitem[Solin et~al.(2016)Solin, S\"arkk\"a, Kannala, and
  Rahtu]{Solin+Sarkka+Kannala+Rahtu:2016}
A.~Solin, S.~S\"arkk\"a, J.~Kannala, and E.~Rahtu, ``Terrain navigation in the
  magnetic landscape: {P}article filtering for indoor positioning,'' in
  \emph{Proceedings of the European Navigation Conference}, 2016.

\bibitem[Woodman(2010)]{Woodman:2010}
O.~J. Woodman, ``Pedestrian localisation for indoor environments,'' Ph.D.
  dissertation, University of Cambridge, United Kingdom, 2010.

\bibitem[Hol(2011)]{Hol:2011}
J.~D. Hol, ``Sensor fusion and calibration of inertial sensors, vision,
  ultra-wideband and {GPS},'' Ph.D. dissertation, Link{\"o}ping University,
  Sweden, 2011.

\bibitem[Leonard and Durrant-Whyte(1991)]{Leonard+Durrant-Whyte:1991}
J.~J. Leonard and H.~F. Durrant-Whyte, ``Simultaneous map building and
  localization for an autonomous mobile robot,'' in \emph{Proceedings of the
  International Workshop of Intelligent Robots and Systems (IROS)},
  vol.~3.\hskip 1em plus 0.5em minus 0.4em\relax IEEE, 1991, pp. 1442--1447.

\bibitem[Durrant-Whyte and Bailey(2006)]{Durrant-Whyte+Bailey:2006}
H.~Durrant-Whyte and T.~Bailey, ``Simultaneous localization and mapping:
  {P}art~{I},'' \emph{IEEE Robotics \& Automation Magazine}, vol.~13, no.~2,
  pp. 99--110, 2006.

\bibitem[O'Hagan(1978)]{OHagan:1978}
A.~O'Hagan, ``Curve fitting and optimal design for prediction (with
  discussion),'' \emph{Journal of the Royal Statistical Society. Series~B
  (Methodological)}, vol.~40, no.~1, pp. 1--42, 1978.

\bibitem[Rasmussen and Williams(2006)]{Rasmussen+Williams:2006}
C.~E. Rasmussen and C.~K.~I. Williams, \emph{Gaussian Processes for Machine
  Learning}.\hskip 1em plus 0.5em minus 0.4em\relax MIT Press, 2006.

\bibitem[Cressie(1993)]{Cressie:1993}
N.~A. Cressie, \emph{Statistics for Spatial Data}.\hskip 1em plus 0.5em minus
  0.4em\relax New York: Wiley-Interscience, 1993.

\bibitem[Cressie and Wikle(2011)]{Cressie+Wikle:2011}
N.~Cressie and C.~K. Wikle, \emph{Statistics for Spatio-Temporal Data}.\hskip
  1em plus 0.5em minus 0.4em\relax Hoboken, NJ: John Wiley \& Sons, 2011.

\bibitem[Deisenroth et~al.(2015)Deisenroth, Fox, and
  Rasmussen]{Deisenroth+Fox+Rasmussen:2015}
M.~P. Deisenroth, D.~Fox, and C.~E. Rasmussen, ``Gaussian processes for
  data-efficient learning in robotics and control,'' \emph{IEEE Transactions on
  Pattern Analysis and Machine Intelligence}, vol.~37, no.~2, pp. 408--423,
  2015.

\bibitem[Wahlstr\"om et~al.(2013)Wahlstr\"om, Kok, Sch\"on, and
  Gustafsson]{Wahlstrom+Kok+Schon+Gustafsson:2013}
N.~Wahlstr\"om, M.~Kok, T.~B. Sch\"on, and F.~Gustafsson, ``Modeling magnetic
  fields using {G}aussian processes,'' in \emph{Proceedings of the
  38\textsuperscript{th} International Conference on Acoustics, Speech and
  Signal Processing (ICASSP)}, 2013, pp. 3522--3526.

\bibitem[Solin and S{\"a}rkk{\"a}(2014)]{Solin+Sarkka:submitted}
A.~Solin and S.~S{\"a}rkk{\"a}, ``Hilbert space methods for reduced-rank
  {G}aussian process regression,'' \emph{ArXiv preprint arXiv:1401.5508}, 2014.

\bibitem[S\"arkk\"a et~al.(2013)S\"arkk\"a, Solin, and
  Hartikainen]{Sarkka+Solin+Hartikainen:2013}
S.~S\"arkk\"a, A.~Solin, and J.~Hartikainen, ``Spatiotemporal learning via
  infinite-dimensional {B}ayesian filtering and smoothing,'' \emph{IEEE Signal
  Processing Magazine}, vol.~30, no.~4, pp. 51--61, 2013.

\bibitem[Nabighian et~al.(2005)Nabighian, Grauch, Hansen, LaFehr, Li, Peirce,
  Phillips, and
  Ruder]{Nabighian+Grauch+Hansen+LaFehr+Li+Peirse+Philips+Ruder:2005}
M.~N. Nabighian, V.~J.~S. Grauch, R.~O. Hansen, T.~R. LaFehr, Y.~Li, J.~W.
  Peirce, J.~D. Phillips, and M.~E. Ruder, ``The historical development of the
  magnetic method in exploration,'' \emph{Geophysics}, vol.~70, no.~6, pp.
  33ND--61ND, 2005.

\bibitem[Guillen et~al.(2008)Guillen, Calcagno, Courrioux, Joly, and
  Ledru]{Guillen+Calcagno+Courrioux+Joly+Ledru:2008}
A.~Guillen, P.~Calcagno, G.~Courrioux, A.~Joly, and P.~Ledru, ``Geological
  modelling from field data and geological knowledge: {P}art {II}. {M}odelling
  validation using gravity and magnetic data inversion,'' \emph{Physics of the
  Earth and Planetary Interiors}, vol. 171, no.~1, pp. 158--169, 2008.

\bibitem[Calcagno et~al.(2008)Calcagno, Chil{\`e}s, Courrioux, and
  Guillen]{Calcagno+Chiles+Courrioux+Guillen:2008}
P.~Calcagno, J.-P. Chil{\`e}s, G.~Courrioux, and A.~Guillen, ``Geological
  modelling from field data and geological knowledge: {P}art {I}. {M}odelling
  method coupling {3D} potential-field interpolation and geological rules,''
  \emph{Physics of the Earth and Planetary Interiors}, vol. 171, no.~1, pp.
  147--157, 2008.

\bibitem[Mackay et~al.(2006)Mackay, Marchand, and
  Kabin]{Mackay+Marchand+Kabin:2006}
F.~Mackay, R.~Marchand, and K.~Kabin, ``Divergence-free magnetic field
  interpolation and charged particle trajectory integration,'' \emph{Journal of
  Geophysical Research: Space Physics}, vol. 111, no.~A6, p. A06208, 2006.

\bibitem[Bhattacharyya(1969)]{Bhattacharyya:1969}
B.~K. Bhattacharyya, ``Bicubic spline interpolation as a method for treatment
  of potential field data,'' \emph{Geophysics}, vol.~34, no.~3, pp. 402--423,
  1969.

\bibitem[Springel(2010)]{Springel:2010}
V.~Springel, ``Smoothed particle hydrodynamics in astrophysics,'' \emph{Annual
  Review of Astronomy and Astrophysics}, vol.~48, pp. 391--430, 2010.

\bibitem[Haverinen and Kemppainen(2009)]{Haverinen+Kemppainen:2009}
J.~Haverinen and A.~Kemppainen, ``Global indoor self-localization based on the
  ambient magnetic field,'' \emph{Robotics and Autonomous Systems}, vol.~57,
  no.~10, pp. 1028--1035, 2009.

\bibitem[Li et~al.(2012)Li, Gallagher, Dempster, and
  Rizos]{Li+Gallagher+Dempster+Rizos:2012}
B.~Li, T.~Gallagher, A.~G. Dempster, and C.~Rizos, ``How feasible is the use of
  magnetic field alone for indoor positioning?'' in \emph{Proceedings of the
  International Conference on Indoor Positioning and Indoor Navigation (IPIN)},
  2012, pp. 1--9.

\bibitem[Angermann et~al.(2012)Angermann, Frassl, Doniec, Julian, and
  Robertson]{Angermann+Frassl+Doniec+Julian+Robertson:2012}
M.~Angermann, M.~Frassl, M.~Doniec, B.~J. Julian, and P.~Robertson,
  ``Characterization of the indoor magnetic field for applications in
  localization and mapping,'' in \emph{Proceedings of the International
  Conference on Indoor Positioning and Indoor Navigation (IPIN)}, 2012, pp.
  1--9.

\bibitem[Le~Grand and Thrun(2012)]{LeGrand+Thrun:2012}
E.~Le~Grand and S.~Thrun, ``3-axis magnetic field mapping and fusion for indoor
  localization,'' in \emph{IEEE Conference on Multisensor Fusion and
  Integration for Intelligent Systems (MFI)}, 2012, pp. 358--364.

\bibitem[Robertson et~al.(2013)Robertson, Frassl, Angermann, Doniec, Julian,
  Garcia~Puyol, Khider, Lichtenstern, and
  Bruno]{Robertson+Frassl+Angermann+Doniec+Julian+Puyol+Khider+Lichtenstern+Bruno:2013}
P.~Robertson, M.~Frassl, M.~Angermann, M.~Doniec, B.~J. Julian,
  M.~Garcia~Puyol, M.~Khider, M.~Lichtenstern, and L.~Bruno, ``Simultaneous
  localization and mapping for pedestrians using distortions of the local
  magnetic field intensity in large indoor environments,'' in \emph{Proceedings
  of the International Conference on Indoor Positioning and Indoor Navigation
  (IPIN)}.\hskip 1em plus 0.5em minus 0.4em\relax IEEE, 2013, pp. 1--10.

\bibitem[Frassl et~al.(2013)Frassl, Angermann, Lichtenstern, Robertson, Julian,
  and Doniec]{Frassl+Angermann+Lichtenstern+Robertson+Julian+Doniec:2013}
M.~Frassl, M.~Angermann, M.~Lichtenstern, P.~Robertson, B.~J. Julian, and
  M.~Doniec, ``Magnetic maps of indoor environments for precise localization of
  legged and non-legged locomotion,'' in \emph{Proceedings of the IEEE/RSJ
  International Conference on Intelligent Robots and Systems (IROS)}, 2013, pp.
  913--920.

\bibitem[Vallivaara et~al.(2010)Vallivaara, Haverinen, Kemppainen, and
  R{\"o}ning]{Vallivaara+Haverinen+Kemppainen+Roning:2010}
I.~Vallivaara, J.~Haverinen, A.~Kemppainen, and J.~R{\"o}ning, ``Simultaneous
  localization and mapping using ambient magnetic field,'' in \emph{Proceedings
  of the IEEE Conference on Multisensor Fusion and Integration for Intelligent
  Systems (MFI)}, 2010, pp. 14--19.

\bibitem[Vallivaara et~al.(2011)Vallivaara, Haverinen, Kemppainen, and
  R{\"o}ning]{Vallivaara+Haverinen+Kemppainen+Roning:2011}
------, ``Magnetic field-based {SLAM} method for solving the localization
  problem in mobile robot floor-cleaning task,'' in \emph{Proceedings of the
  15\textsuperscript{th} International Conference on Advanced Robotics (ICAR)},
  2011, pp. 198--203.

\bibitem[O'Callaghan and Ramos(2012)]{OCallaghan+Ramos:2012}
S.~T. O'Callaghan and F.~T. Ramos, ``Gaussian process occupancy maps,''
  \emph{The International Journal of Robotics Research}, vol.~31, no.~1, pp.
  42--62, 2012.

\bibitem[Smith et~al.(2011)Smith, Posner, and Newman]{Smith+Posner+Newman:2011}
M.~Smith, I.~Posner, and P.~Newman, ``Adaptive compression for {3D} laser
  data,'' \emph{The International Journal of Robotics Research}, vol.~30,
  no.~7, pp. 914--935, 2011.

\bibitem[Kim and Kim(2015)]{Kim+Kim:2015}
S.~Kim and J.~Kim, ``Hierarchical {G}aussian processes for robust and accurate
  map building,'' in \emph{Proceedings of Australasian Conference on Robotics
  and Automation}, 2015, pp. 117--124.

\bibitem[Ramos and Ott(2016)]{Ramos+Ott:2016}
F.~Ramos and L.~Ott, ``Hilbert maps: {S}calable continuous occupancy mapping
  with stochastic gradient descent,'' \emph{The International Journal of
  Robotics Research}, vol.~35, no.~14, pp. 1717--1730, 2016.

\bibitem[Senanayake et~al.(2016)Senanayake, Ott, O'Callaghan, and
  Ramos]{Senanayake+Ott+Callaghan+Ramos:2016}
R.~Senanayake, L.~Ott, S.~O'Callaghan, and F.~T. Ramos, ``Spatio-temporal
  {H}ilbert maps for continuous occupancy representation in dynamic
  environments,'' in \emph{Advances in Neural Information Processing Systems
  29}.\hskip 1em plus 0.5em minus 0.4em\relax Curran Associates, Inc., 2016,
  pp. 3925--3933.

\bibitem[Vidal-Calleja et~al.(2014)Vidal-Calleja, Su, De~Bruijn, and
  Miro]{Vidal-Calleja+Su+Bruijn+Miro:2014}
T.~Vidal-Calleja, D.~Su, F.~De~Bruijn, and J.~V. Miro, ``Learning spatial
  correlations for {B}ayesian fusion in pipe thickness mapping,'' in
  \emph{Proceedings of the IEEE International Conference on Robotics and
  Automation (ICRA)}, 2014, pp. 683--690.

\bibitem[Tong et~al.(2013)Tong, Furgale, and
  Barfoot]{Tong+Furgale+Barfoot:2013}
C.~H. Tong, P.~Furgale, and T.~D. Barfoot, ``Gaussian process {G}auss--{N}ewton
  for non-parametric simultaneous localization and mapping,'' \emph{The
  International Journal of Robotics Research}, vol.~32, no.~5, pp. 507--525,
  2013.

\bibitem[Ferris et~al.(2006)Ferris, H{\"a}hnel, and
  Fox]{Ferris+Hahnel+Fox:2006}
B.~Ferris, D.~H{\"a}hnel, and D.~Fox, ``Gaussian processes for signal
  strength-based location estimation,'' in \emph{Proceedings of Robotics:
  Science and Systems}, 2006.

\bibitem[Barkby et~al.(2012)Barkby, Williams, Pizarro, and
  Jakuba]{Barkby+Williams+Pizarro+Jakuba:2012}
S.~Barkby, S.~B. Williams, O.~Pizarro, and M.~V. Jakuba, ``Bathymetric particle
  filter {SLAM} using trajectory maps,'' \emph{The International Journal of
  Robotics Research}, vol.~31, no.~12, pp. 1409--1430, 2012.

\bibitem[Barfoot et~al.(2014)Barfoot, Tong, and
  S\"arkk\"a]{Barfoot+Tong+Sarkka:2014}
T.~D. Barfoot, C.~H. Tong, and S.~S\"arkk\"a, ``Batch continuous-time
  trajectory estimation as exactly sparse {G}aussian process regression,'' in
  \emph{Proceedings of Robotics: Science and Systems}, 2014.

\bibitem[Anderson et~al.(2015)Anderson, Barfoot, Tong, and
  S\"arkk\"a]{Anderson+Barfoot+Tong+Sarkka:2015}
S.~Anderson, T.~D. Barfoot, C.~H. Tong, and S.~S\"arkk\"a, ``Batch nonlinear
  continuous-time trajectory estimation as exactly sparse {G}aussian process
  regression,'' \emph{Autonomous Robots}, vol.~39, no.~3, pp. 221--238, 2015.

\bibitem[Qui{\~n}onero-Candela and
  Rasmussen(2005)]{Quinonero-Candela+Rasmussen:2005}
J.~Qui{\~n}onero-Candela and C.~E. Rasmussen, ``A unifying view of sparse
  approximate {G}aussian process regression,'' \emph{Journal of Machine
  Learning Research}, vol.~6, pp. 1939--1959, 2005.

\bibitem[L{\'a}zaro-Gredilla et~al.(2010)L{\'a}zaro-Gredilla,
  Qui{\~n}onero-Candela, Rasmussen, and
  Figueiras-Vidal]{Lazaro-Gredilla+Quinonero-Candela+Rasmussen+Figueiras-Vidal:2010}
M.~L{\'a}zaro-Gredilla, J.~Qui{\~n}onero-Candela, C.~E. Rasmussen, and A.~R.
  Figueiras-Vidal, ``Sparse spectrum {G}aussian process regression,''
  \emph{Journal of Machine Learning Research}, vol.~11, pp. 1865--1881, 2010.

\bibitem[Hensman et~al.(accepted)Hensman, Durrande, and
  Solin]{Hensman+Durrande+Solin:2018}
J.~Hensman, N.~Durrande, and A.~Solin, ``Variational {F}ourier features for
  {G}aussian processes,'' \emph{Journal of Machine Learning Research},
  accepted.

\bibitem[Paciorek(2007)]{Paciorek:2007}
C.~J. Paciorek, ``Bayesian smoothing with {G}aussian processes using {F}ourier
  basis functions in the {spectralGP} package,'' \emph{Journal of Statistical
  Software}, vol.~19, no.~2, pp. 1--38, 2007.

\bibitem[Fritz et~al.(2009)Fritz, Neuweiler, and
  Nowak]{Fritz+Neuweiler+Nowak:2009}
J.~Fritz, I.~Neuweiler, and W.~Nowak, ``Application of {FFT}-based algorithms
  for large-scale universal kriging problems,'' \emph{Mathematical
  Geosciences}, vol.~41, no.~5, pp. 509--533, 2009.

\bibitem[Hartikainen and S\"arkk\"a(2010)]{Hartikainen+Sarkka:2010}
J.~Hartikainen and S.~S\"arkk\"a, ``Kalman filtering and smoothing solutions to
  temporal {G}aussian process regression models,'' in \emph{Proceedings of the
  International Workshop on Machine Learning for Signal Processing
  (MLSP)}.\hskip 1em plus 0.5em minus 0.4em\relax IEEE, 2010, pp. 379--384.

\bibitem[Reece and Roberts(2010)]{Reece+Roberts:2010}
S.~Reece and S.~Roberts, ``An introduction to {G}aussian processes for the
  {K}alman filter expert,'' in \emph{Proceedings of the 13th International
  Conference on Information Fusion ({FUSION})}, 2010, pp. 1--9.

\bibitem[Osborne(2010)]{Osborne:2010}
M.~Osborne, ``Bayesian {G}aussian processes for sequential prediction,
  optimisation and quadrature,'' Ph.D. dissertation, University of Oxford,
  United Kingdom, 2010.

\bibitem[Huber(2014)]{Huber:2014}
M.~F. Huber, ``Recursive {G}aussian process: {O}n-line regression and
  learning,'' \emph{Pattern Recognition Letters}, vol.~45, pp. 85--91, 2014.

\bibitem[\'{A}lvarez et~al.(2013)\'{A}lvarez, Luengo, and
  Lawrence]{Alvarez+Luengo+Lawrence:2013}
M.~A. \'{A}lvarez, D.~Luengo, and N.~D. Lawrence, ``Linear latent force models
  using {G}aussian processes,'' \emph{IEEE Transactions on Pattern Analysis and
  Machine Intelligence}, vol.~35, no.~11, pp. 2693--2705, 2013.

\bibitem[\'{A}lvarez and Lawrence(2009)]{Alvarez+Lawrence:2009}
M.~A. \'{A}lvarez and N.~D. Lawrence, ``Latent force models,'' in
  \emph{Proceedings of the 12\textsuperscript{th} International Conference on
  Artificial Intelligence and Statistics}, ser. JMLR W\&CP, vol.~5, 2009, pp.
  9--16.

\bibitem[Hartikainen and S\"arkk\"a(2011)]{Hartikainen+Sarkka:2011}
J.~Hartikainen and S.~S\"arkk\"a, ``Sequential inference for latent force
  models,'' in \emph{Proceedings of the International Conference on Uncertainty
  in Artificial Intelligence}, 2011, pp. 311--318.

\bibitem[S\"arkk\"a and Hartikainen(2012)]{Sarkka+Hartikainen:2012}
S.~S\"arkk\"a and J.~Hartikainen, ``Infinite-dimensional {K}alman filtering
  approach to spatio-temporal {G}aussian process regression,'' in
  \emph{Proceedings of the 15\textsuperscript{th} International Conference on
  Artificial Intelligence and Statistics}, ser. JMLR W\&CP, vol.~22, 2012, pp.
  993--1001.

\bibitem[Curtain and Pritchard(1978)]{Curtain+Pritchard:1978}
R.~F. Curtain and A.~J. Pritchard, \emph{Infinite Dimensional Linear Systems
  Theory}.\hskip 1em plus 0.5em minus 0.4em\relax Springer--Verlag, 1978.

\bibitem[Jackson(1999)]{Jackson:1999}
J.~D. Jackson, \emph{Classical Electrodynamics}, 3rd~ed.\hskip 1em plus 0.5em
  minus 0.4em\relax New York: Wiley, 1999.

\bibitem[Vanderlinde(2004)]{Vanderlinde:2004}
J.~Vanderlinde, \emph{Classical Electromagnetic Theory}.\hskip 1em plus 0.5em
  minus 0.4em\relax Dordrecht: Springer, 2004.

\bibitem[Griffiths and College(1999)]{Griffiths+College:1999}
D.~J. Griffiths and R.~College, \emph{Introduction to Electrodynamics}.\hskip
  1em plus 0.5em minus 0.4em\relax Upper Saddle River, NJ: Prentice hall, 1999.

\bibitem[Wahlstr\"om(2015)]{Wahlstrom:2015}
N.~Wahlstr\"om, ``Modeling of magnetic fields and extended objects for
  localization applications,'' Ph.D. dissertation, Link\"oping University,
  Sweden, 2015.

\bibitem[Kemppainen et~al.(2011)Kemppainen, Haverinen, Vallivaara, and
  R{\"o}ning]{Kemppainen+Haverinen+Vallivaara+Roning:2011}
A.~Kemppainen, J.~Haverinen, I.~Vallivaara, and J.~R{\"o}ning, ``Near-optimal
  exploration in {G}aussian process {SLAM}: {S}calable optimality factor and
  model quality rating,'' in \emph{Proceedings of the 5\textsuperscript{th}
  European Conference on Mobile Robots (ECMR)}, 2011, pp. 283--289.

\bibitem[Jung et~al.(2015)Jung, Oh, and Myung]{Jung+Oh+Myung:2015}
J.~Jung, T.~Oh, and H.~Myung, ``Magnetic field constraints and sequence-based
  matching for indoor pose graph {SLAM},'' \emph{Robotics and Autonomous
  Systems}, vol.~70, pp. 92--105, 2015.

\bibitem[Viseras~Ruiz and Olariu(2015)]{Ruiz+Olariu:2015}
A.~Viseras~Ruiz and C.~Olariu, ``A general algorithm for exploration with
  {G}aussian processes in complex, unknown environments,'' in \emph{Proceeding
  of the IEEE International Conference on Robotics and Automation (ICRA)},
  2015, pp. 3388--3393.

\bibitem[Fuselier(2007)]{Fuselier:2007}
E.~J. Fuselier, Jr, ``Refined error estimates for matrix-valued radial basis
  functions,'' Ph.D. dissertation, Texas A\&M University, 2007.

\bibitem[Baldassarre et~al.(2010)Baldassarre, Rosasco, Barla, and
  Verri]{Baldassarre+Rosasco+Barla+Verri:2010}
L.~Baldassarre, L.~Rosasco, A.~Barla, and A.~Verri, ``Vector field learning via
  spectral filtering,'' in \emph{Machine Learning and Knowledge Discovery in
  Databases}, ser. Lecture Notes in Computer Science.\hskip 1em plus 0.5em
  minus 0.4em\relax Springer, 2010, vol. 6321, pp. 56--71.

\bibitem[\'{A}lvarez et~al.(2012)\'{A}lvarez, Rosasco, and
  Lawrence]{Alvarez+Rosasco+Lawrence:2012}
M.~A. \'{A}lvarez, L.~Rosasco, and N.~D. Lawrence, ``Kernels for vector-valued
  functions: {A} review,'' \emph{Foundations and Trends in Machine Learning},
  vol.~4, no.~3, pp. 195--266, 2012.

\bibitem[Jidling et~al.(2017)Jidling, Wahlstr{\"o}m, Wills, and
  Sch{\"o}n]{Jidling+Wahlstrom+Wills+Schon:2017}
C.~Jidling, N.~Wahlstr{\"o}m, A.~Wills, and T.~B. Sch{\"o}n, ``Linearly
  constrained {G}aussian processes,'' in \emph{Advances in Neural Information
  Processing Systems (NIPS) 30}, 2017.

\bibitem[S{\"a}rkk{\"a}(2013)]{Sarkka:2013}
S.~S{\"a}rkk{\"a}, \emph{Bayesian Filtering and Smoothing}.\hskip 1em plus
  0.5em minus 0.4em\relax Cambridge University Press, 2013.

\bibitem[Tarantola(2004)]{Tarantola:2004}
A.~Tarantola, \emph{Inverse Problem Theory and Methods for Model Parameter
  Estimation}.\hskip 1em plus 0.5em minus 0.4em\relax SIAM, 2004.

\bibitem[Kaipio and Somersalo(2005)]{Kaipio+Somersalo:2005}
J.~Kaipio and E.~Somersalo, \emph{Statistical and Computational Inverse
  Problems}.\hskip 1em plus 0.5em minus 0.4em\relax Springer, 2005.

\bibitem[S\"arkk\"a(2011)]{Sarkka:2011b}
S.~S\"arkk\"a, ``Linear operators and stochastic partial differential equations
  in {G}aussian process regression,'' in \emph{Artificial Neural Networks and
  Machine Learning -- {ICANN} 2011}, ser. Lecture Notes in Computer Science,
  vol. 6792.\hskip 1em plus 0.5em minus 0.4em\relax Springer, 2011, pp.
  151--158.

\end{thebibliography}
}

\begin{IEEEbiography}[{%
  \includegraphics[width=1in,height=1.25in,clip,keepaspectratio]{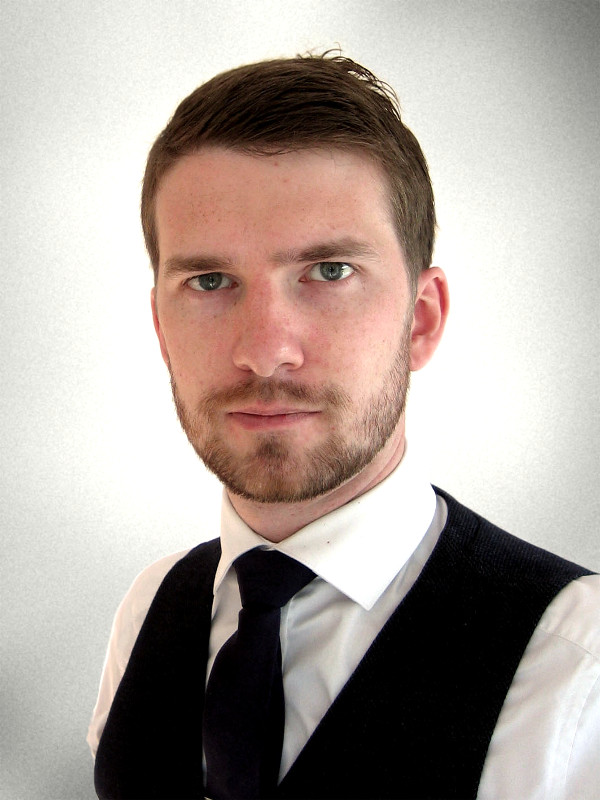}}]{Arno Solin} received his M.Sc.\ (Tech.) degree (with distinction) in Engineering Physics and Mathematics and D.Sc.\ (Tech.) degree (with distinction) in computational science from Aalto University, Finland, in 2012 and 2016, respectively. 

  Dr.\ Solin is an Academy of Finland Postdoctoral Researcher with Aalto University, and Technical Advisor of IndoorAtlas Ltd. His research interests are in probabilistic inference for temporal, spatial, and spatio-temporal models with applications in sensor fusion for tracking and navigation, brain imaging, and machine learning problems.
\end{IEEEbiography}

\vspace*{-1em}

\begin{IEEEbiography}[{%
  \includegraphics[width=1in,height=1.25in,clip,keepaspectratio]{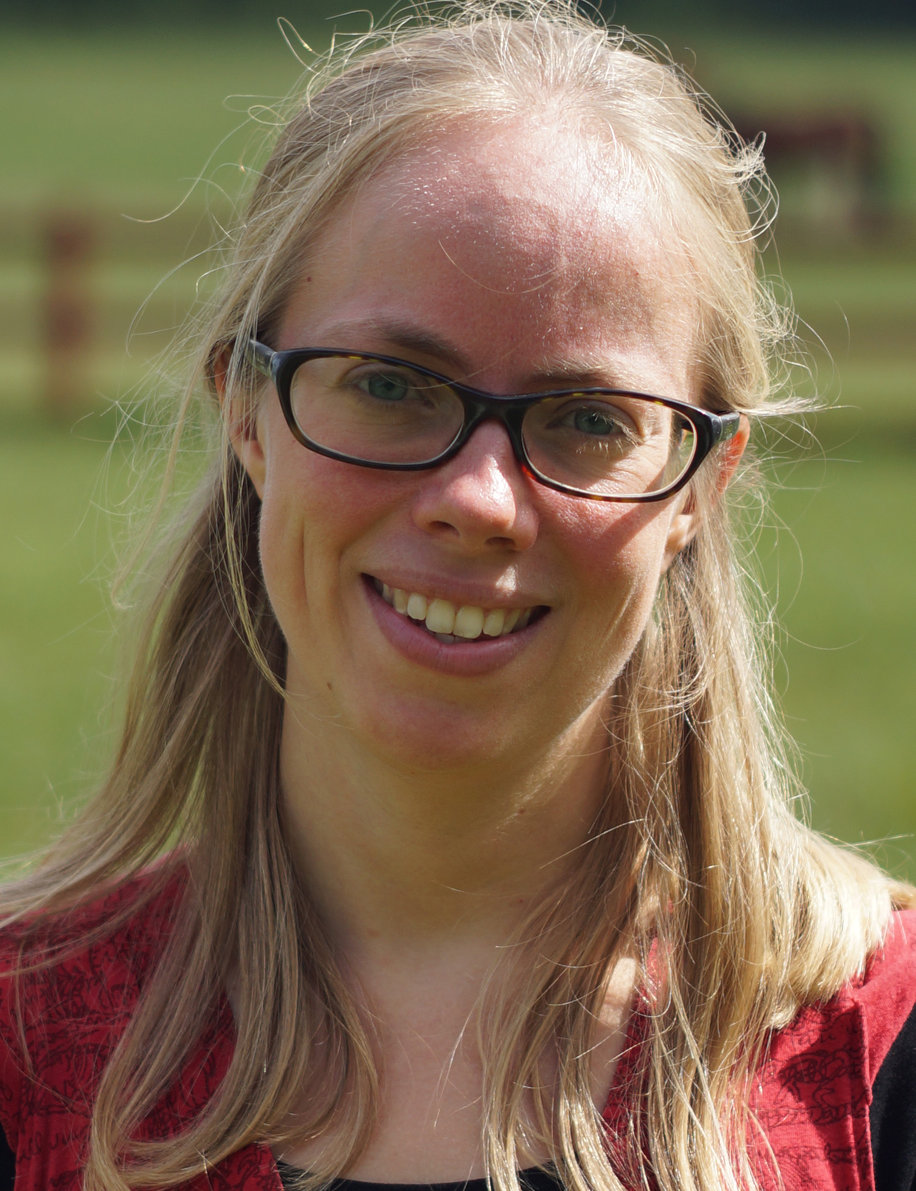}}]{Manon Kok} received M.Sc.\ degrees in Applied Physics and in Philosophy of Science, Technology and Society (2009 and 2007, respectively), both from the University of Twente. From 2009 until 2011 she worked as a Research Engineer at Xsens Technologies. She received the PhD degree in Automatic Control from Link\"oping University in 2017. 

  Dr.\ Kok is currently a Postdoc in the Machine Learning Group of the Computational and Biological Learning Lab at the University of Cambridge. Her research interests are in the fields of probabilistic inference for sensor fusion, signal processing and machine learning. 
\end{IEEEbiography}

\vspace*{-1em}

\begin{IEEEbiography}[{%
  \includegraphics[width=1in,height=1.25in,clip,keepaspectratio]{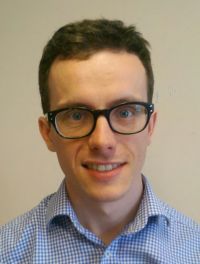}}]{Niklas Wahlstr\"om} received the M.Sc.\ degree in Applied Physics and Electrical Engineering in 2010 and Ph.D.\ degree in automatic control in 2015, both from Link\"oping University, Sweden.
	
  Dr. Wahlstr\"om is since 2016 a Researcher at the Department of Information Technology, Uppsala University. His research interests include machine learning, deep learning, sensor fusion, localization and mapping, especially using magnetic sensors.

\end{IEEEbiography}

\vspace*{-1em}

\begin{IEEEbiography}[{%
  \includegraphics[width=1in,height=1.25in,clip,keepaspectratio]{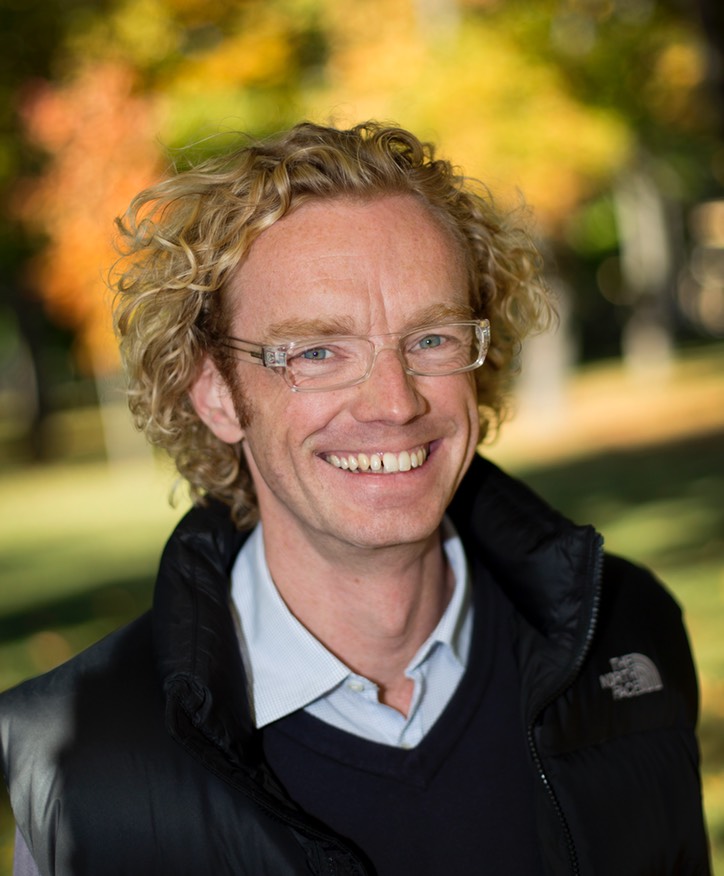}}]{Thomas B.\ Sch\"on} received the M.Sc.\ degree in Applied Physics and Electrical Engineering in 2001, the B.Sc.\ degree in Business Administration 2001 and the Ph.D. degree in Automatic Control in 2006, all from Link\"oping University.

  Dr.\ Sch\"on is Professor of the Chair of Automatic Control in the Department of Information Technology at Uppsala University. His main research interest is nonlinear inference problems, especially within the context of dynamical systems, solved using probabilistic methods, more specifically sequential Monte Carlo, particle MCMC and graphical models. He is active within the fields of machine learning, signal processing and automatic control. He is a Senior Member of the IEEE.
\end{IEEEbiography}

\vspace*{-1em}

\begin{IEEEbiography}[{%
  \includegraphics[width=1in,height=1.25in,clip,keepaspectratio]{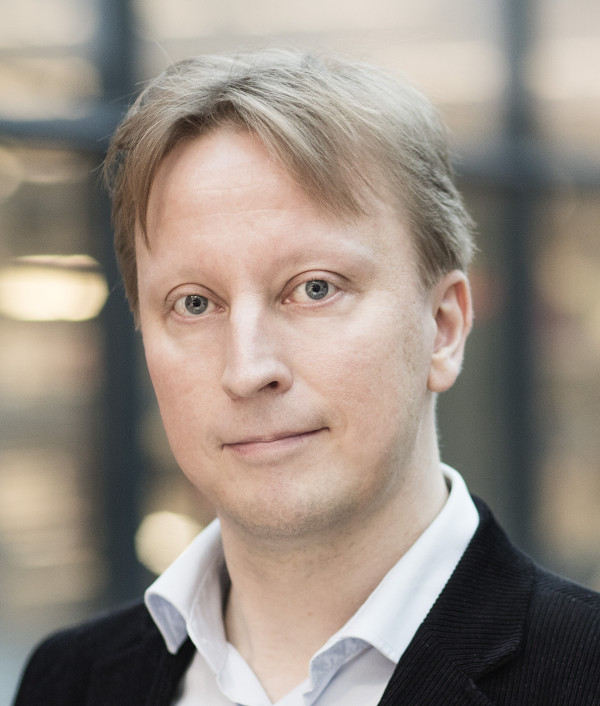}}]{Simo S\"arkk\"a}
received his M.Sc.\ (Tech.) degree (with distinction) in Engineering Physics and Mathematics, and D.Sc.\ (Tech.) degree (with distinction) in electrical and communications engineering from Helsinki University of Technology, Espoo, Finland, in 2000 and 2006, respectively.

  Dr.\ S\"arkk\"a is an Associate Professor and Academy Research Fellow with Aalto University, Technical Advisor and Director of IndoorAtlas Ltd., and an Adjunct Professor with Tampere University of Technology and Lappeenranta University of Technology. His research interests are in multi-sensor data processing systems with applications in location sensing, health technology, machine learning, inverse problems, and brain imaging. He is a Senior Member of the IEEE.
\end{IEEEbiography}

\vspace*{\fill}

\end{document}